\documentclass{article} 
\usepackage[vmargin=1.18in,hmargin=1.1in]{geometry}
\usepackage{amsmath,amssymb,amsfonts,graphicx,amsthm,nicefrac,mathtools,bm}

\usepackage{hyperref} 
\usepackage[numbers]{natbib}
\usepackage[english]{babel}
\usepackage{times}
\usepackage[T1]{fontenc}
\usepackage{bbm,mdframed}
\usepackage[dvipsnames]{xcolor}
\usepackage{xspace}
\usepackage{rotating,multirow} 
\usepackage{tikz}
\usetikzlibrary{arrows}
\usetikzlibrary{shapes} 
\usepackage{algpseudocode,algorithm,algorithmicx} 
\usepackage{fancyhdr} 
\usepackage{tabularx}
\usepackage{comment}
\usepackage{transparent}
\usepackage{makecell}
\usepackage{longtable} 
\usepackage{subcaption}
\usepackage{enumitem}

\date{}
\author{\theauthor}
\title{\thetitle} 
\fancypagestyle{plain}{%
  \fancyhf{}%
  \lhead{\thepage}
}   

\newcommand{\theauthor}{}
\newcommand{\thetitle}{Greedy Attack and Gumbel Attack:\\ Generating Adversarial Examples for Discrete Data}

\newcommand{\Prob}{\ensuremath{\mathbb{P}}}

\newcommand{\Y}{\mathcal{Y}}
\newcommand{\dict}{\mathbb W}
\newcommand{\Exp}{\ensuremath{\mathbb{E}}}

\newcommand{\real}{\ensuremath{\mathbb{R}}}
\newcommand{\ProbModel}{\ensuremath{\Prob_m}}

\newcommand{\defn}{\ensuremath{: \, = }}

\newcommand{\mypara}[1]{\paragraph{#1.}}

\begin{document}

\author{
  \begin{tabular}[t]{c}
  Puyudi Yang\thanks{* indicates equal contribution.} \\ 
  University of California, Davis\\ 
  \texttt{pydyang@ucdavis.edu} 
  \end{tabular}
   \begin{tabular}[t]{c}
  Jianbo Chen$^*$ \\
  University of California, Berkeley \\
  \texttt{jianbochen@berkeley.edu}\\ 
  \end{tabular}\\
  \\
  \begin{tabular}[t]{c}
  Cho-Jui Hsieh\\
  University of California, Davis \\
  \texttt{chohsieh@ucdavis.edu} 
  \end{tabular} 
   \begin{tabular}[t]{c}
  Jane-Ling Wang \\
  University of California, Davis \\
  \texttt{janelwang@ucdavis.edu} 
  \end{tabular} \\
  \\
  \begin{tabular}[t]{c}
  Michael I. Jordan \\
  University of California, Berkeley \\
  \texttt{jordan@cs.berkeley.edu} 
  \end{tabular}
}

\maketitle

\begin{abstract} 
We present a probabilistic framework for studying adversarial attacks on discrete data. Based on this framework, we derive a perturbation-based method, \emph{Greedy Attack}, and a scalable learning-based method, \emph{Gumbel Attack}, that illustrate various tradeoffs in the design of attacks. We demonstrate the effectiveness of these methods using both quantitative metrics and human evaluation on various state-of-the-art models for text classification, including a word-based CNN, a character-based CNN and an LSTM. As as example of our results, we show that the accuracy of character-based convolutional networks drops to the level of random selection by modifying only five characters through Greedy Attack.   
\end{abstract}

\section{Introduction}
Robustness to adversarial perturbation has become an extremely important criterion for applications of machine learning in security-sensitive domains such as spam detection \cite{stringhini2010detecting}, fraud detection \cite{ghosh1994credit}, criminal justice \cite{berk2013statistical}, malware detection \cite{kolter2006learning}, and financial markets \cite{west2000neural}. 
Systematic methods for generating adversarial examples by small perturbations of original input data, also known as ``attack,'' 
have been developed to operationalize this criterion and to drive the development of more robust learning systems~\cite{dalvi2004adversarial, szegedy2013intriguing, goodfellow2014explaining}.

Most of the work in this area has focused on differentiable models with continuous input spaces~\cite{szegedy2013intriguing, goodfellow2014explaining, kurakin2016adversarial, kurakin2016adversarial}. In this setting, the proposed attack strategies add a gradient-based perturbation to the original input.  It has been shown that such perturbations can result in a dramatic decrease in the predictive accuracy of the model. Thus this line of research has demonstrated the vulnerability of deep neural networks to adversarial examples in tasks like image classification and speech recognition.

We focus instead on adversarial attacks on models with discrete input data, such as text data, where each feature of an input sample has a categorical domain.  While gradient-based approaches are not directly applicable to this setting, variations of gradient-based approaches have been shown effective in differentiable models.  For example, \citet{li2015visualizing} proposed to locate the top features with the largest gradient magnitude of their embedding, and \citet{papernot2016crafting} proposed to modify randomly selected features of an input through perturbing each feature by signs of the gradient, and project them onto the closest vector in the embedding space. \citet{dalvi2004adversarial} attacked such models by solving an integer linear program. \citet{gao2018black} developed scoring functions applicable for sequence data, and proposed to modify characters of the features selected by the scoring functions. Attack methods specifically designed for text data have also been studied recently. \citet{jia2017adversarial} proposed to insert distraction sentences into samples in a human-involved loop to fool a reading comprehension system. \citet{samanta2017towards} added linguistic constraints over the pool of candidate-replacing words.

We propose a two-stage probabilistic framework for generating adversarial examples for models with discrete input, where the key features to be perturbed are identified in the first stage and subsequently perturbed in the second stage by values chosen from a pre-fixed dictionary. We derive two methods---\emph{Greedy Attack} and \emph{Gumbel Attack}---based on the proposed framework. Greedy attack evaluates models with single-feature perturbed inputs in two stages, while Gumbel Attack learns a parametric sampling distribution for perturbation. Greedy Attack achieves higher success rate, while Gumbel Attack requires fewer model evaluations, leading to better efficiency in real-time or large-scale attacks. 
Table~\ref{tab:prop-summary} systematically compares our methods with other methods.  

In summary, our contributions in this work are as follows:
(1) We propose a probabilistic framework for adversarial attacks on models with discrete data. (2) We show that Greedy Attack achieves state-of-the-art attack rates across various kinds of models. (3) We propose Gumbel Attack as a scalable method with low model evaluation complexity.  (4) We observe that character-based models in text classification are particularly vulnerable to adversarial attack.

\begin{table}[t]
\centering  
\resizebox{0.7\linewidth}{!}{
\begin{tabular}{||c|c|c|c|c||}
\hline
&  Training &  Efficiency &  Success rate & Black-box\\
\hline
Saliency \cite{simonyan2013deep, liang2017deep}& No & High & Medium & No \\
Projected FGSM \cite{papernot2016crafting} & No & High & Low & No \\
Delete 1-score \cite{li2016understanding} & No & Low & High & Yes \\
DeepWordBug & No & Low & Medium & Yes\\
Greedy Attack & No & Low & Highest & Yes \\
Gumbel Attack & Yes & High & Medium & Yes / No\\ 
 \hline
\end{tabular}
}

\caption{Methods comparisons. 
``Efficiency'': computational time and model evaluation times. ``Black-box'': applicability to black-box models. See Section~\ref{sec:exp} for details.}
\label{tab:prop-summary}

\end{table}

\begin{figure}[t]

\centering
\includegraphics[width=0.7\linewidth]{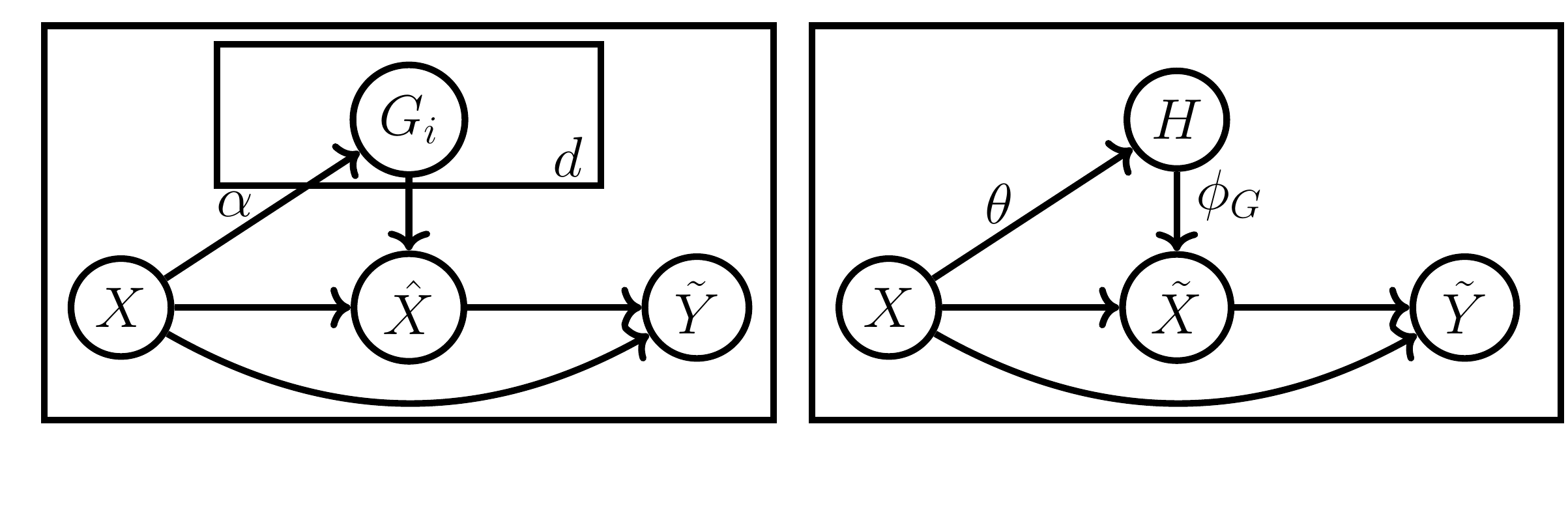}
\vspace{-5mm}
\caption{The left and right figures show the graphical models of the first and second stage respectively.}

\label{fig:model} 
\end{figure}

\section{Framework}
We assume a model in the form of a conditional distribution, $\ProbModel(Y\mid x)$, for a response $Y$, supported on a set $\Y$, given a realization of an input random variable $X=x\in \dict^d$, where $\dict\defn \{w_0,w_1,\dots,w_{m}\}$ is a discrete space such as the dictionary of words, or the space of characters. We assume there exists $w_0\in \dict$ that can be taken as a reference point with no contribution to classification. For example, $w_0$ can be the zero padding in text classification. We define $\tilde Y \mid x,\tilde x \defn \pmb 1 \{\arg\max_y\ProbModel (y\mid \tilde x) \neq \arg\max_y\ProbModel (y\mid x)\}$, as the indicator variable of a successful attack, where $\tilde x$ is the perturbed sample. The goal of the adversarial attack is to turn a given sample $x$ into $\tilde x$ through small perturbations, so that $\tilde Y=1$ given $\tilde x$. We restrict the perturbations to $k$ features of $x$, and approach the problem through two stages. In the first stage, we search for the most important $k$ features of $x$. In the second stage, we search for values to replace the selected $k$ features:
\begin{align}
\label{opt:step1} 
 \text{First stage:}\ \ \hat x = \arg\max_{a\in S_1(x, k)} &\Prob(\tilde Y = 1|a ,x), \\
\label{opt:step2}   
  \text{Second stage:} \ \ \tilde x = \arg \max_{a \in S_2(\hat x, x)} &\Prob(\tilde Y = 1|a,x),
\end{align} 
where $S_1(x, k)\defn\{a\in \dict^d\mid a_i \in \{x_i, w_0\} \text{ for all } i, d(a,x)\leq k\}$ is a set containing all the elements that differ from $x$ by at most $k$ positions, with the different features always taking value $w_0$, and $S_2(\hat x, x)\defn \{a\in \dict^d\mid a_i = \hat x_i \text{ if } \hat x_i = x_i; a_i \in \dict' \text{ otherwise}\}$. Here, we denote by $x_i,a_i, \hat x_i$ the $i$th feature of $x, a, \hat x$, by $d(a,x)$ the count of features different between $a$ and $x$, and by $\dict'\subset \dict$ a sub-dictionary of $\dict$ chosen by the attacker.  

These two objectives are computationally intractable in general. We thus further propose a probabilistic framework to reformulate the objectives in a more tractable manner, as shown in Figure~\ref{fig:model}. Let $G$ be a random variable in $D^d_k:=\{z\in\{0,1\}^d:\sum z_i = k\}$, the space of $d$-dimensional zero-one vectors with $k$ ones, and $\phi:\dict^d\times D^d_k\to \dict^d$ be a function such that $\phi(x,g)_i = x_i$ if $g_i = 0$ and let $\phi(x,g)_i = w_0$ otherwise. In the first stage, we let $\hat X = \phi(X,G)$ where $G$ is generated from a distribution conditioned on $X$. We further add a constraint on $\Prob(G|X)$, by defining $k$ independent and identical random one-hot random variables $G^1,G^2,\dots,G^k\in D^d_1$, and letting $G_i\defn \max_s\{G_i^s\}$, with $G_i$ and $G_i^s$ being the $i$th entries of the variable $G$ and $G^s$ respectively. We aim to maximize the objective $\Prob (\tilde Y = 1\mid \hat X, X)$ over the distribution of $G$ given $X$, the probability of successful attack obtained by merely masking features:
\begin{align}
  \underset{\Prob(G\mid x)}{\text{max}}\Exp_X[\Prob (\tilde Y = 1 \mid \hat X, X)], 
  \text{ s.t. } G^s\overset{i.i.d.}{\sim} \Prob (\cdot\mid X),~ \hat X = \phi(X,G),~\tilde Y\sim \Prob (\tilde Y\mid \hat X,X). \label{opt:prob1} 
\end{align}
The categorical distribution $\Prob(G^s\mid x)$ yields a rank over the $d$ features for a given $x$. We define $\phi^G:\dict^d\to \mathcal P_k([d])$ to be the deterministic function that maps an input $x$ to the indices of the top $k$ features based on the rank from $\Prob(G^s\mid x)$: $\phi^G(x)=\{i_1,\dots,i_k\}$.

In the second stage, we introduce a new random variable $H = (H^{1},\dots, H^{d})$ with each $H^{i}$ being a one-hot random variable in $D^{|\dict'|}_1:=\{z\in\{0,1\}^{|\dict'|}:\sum z_i = 1\}$. Let $\mathcal P_k([d])$ be the set of subsets of $[d]$ of size $k$.  Let $\psi:\dict^d\times (D^{|\dict'|}_1)^d\times \mathcal P_k([d])\to \dict^d$ be a function such that $\psi(x,h,\phi^G(x))_i$ is defined to be $x_i$ if $i\notin \phi^G(x)$, and is the value in $\dict'$ corresponding to the one-hot vector $h_i$ otherwise. The perturbed input is $\tilde X \defn \psi(X,H,\phi^G(X))$, where $H$ is generated from a distribution conditioned on $X$.  We add a constraint on $\Prob (H\mid X)$ by requiring $H^{1},\dots, H^{d}$ to be independent of each other conditioned on $X$. Our goal is to maximize the objective $\Prob (\tilde Y = 1\mid \tilde X, X)$ over the distribution of $H$ given $X$: 
\begin{align} 
  \underset{\Prob (H\mid x)}{\text{max}}\Exp_X[\Prob (\tilde Y = 1 | \tilde X,X)],  
  \text{ s.t. } H\sim \Prob (\cdot | X),~ \tilde X = \psi(X,H,\phi^G(X)),~\tilde Y\sim \Prob (\tilde Y|\tilde X,X).  \label{opt:prob2}
\end{align} 
For a given input $x$, the categorical distribution $\Prob(H^i\mid x)$ yields a rank over the values in $\dict'$ to be chosen for each feature $i$. The perturbation on $x$ is carried out on the top $k$ features $\phi^G(x)=\{i_1,\dots,i_k\}$ ranked by $\Prob(G^s\mid x)$, each chosen feature $i_s$ assigned the top value in $\dict'$ selected by $\Prob(H^{i_s}\mid x)$.

\section{Methods}
In this section we present two instantiations of our general framework: \emph{Greedy Attack} and \emph{Gumbel Attack}.

\subsection{Greedy Attack}
Let $e_i$ denote the $d$-dimensional one-hot vector whose $i$th component is $1$. To solve Problem~\eqref{opt:prob1}, we decompose the objective as:
\vspace{-1mm}
\begingroup\makeatletter\def\f@size{9}\check@mathfonts
\def\maketag@@@#1{\hbox{\m@th\large\normalfont#1}}
\begin{align*}
\Exp[\Prob (\tilde Y = 1 \mid \hat X, X)] &= \Exp_X\big [\sum_{i=1}^d\Prob(G^1=e_i\mid X)\Exp[\Prob(\tilde Y = 1\mid \hat X)\mid X, G^1 = e_i]\big]\nonumber\\
&= \Exp_X\big [\sum_{i=1}^d\Prob(G^1=e_i\mid X)\Exp[\Prob(\tilde Y = 1\mid \phi(X,(e_i,G^2,\dots,G^k)))\mid X]\big ], 
\end{align*}
\endgroup
where the second equality follows from the independence assumption. Given a realization $x$ of the input variable $X$, while the exact value of the quantity $\Exp[\Prob(\tilde Y = 1\mid \phi(x,(e_i,G^2,\dots,G^k)))]$ is hard to compute, a relatively efficient approximation can be constructed by using the output of the model with only the $i$th feature (corresponding to $e_i$) being masked with the reference value $w_0$: 
\begin{equation}
\label{eq:approx1}
\Exp[\Prob(\tilde Y = 1\mid \phi(x,(e_i,G^2,\dots,G^k)))\mid x]\approx 1-\Prob(\tilde Y = 1 \mid x_{(i)}),
\end{equation}
where $x_{(i)}$ replaces the $i$th feature of $x$ with $w_0$. We observe that the approximated objective is maximized if 
\begin{equation}
\Prob(G^1=e_i\mid x)\propto 1 - \Prob(\tilde Y = 1 \mid x_{(i)}).\label{eq:loo1}
\end{equation}


\noindent Similarly, we decompose the objective in Problem~\eqref{opt:prob2} by conditioning on $H^{i_1}$, and again use a greedy approximation:
\begin{align}
&\Exp_X[\Prob (\tilde Y = 1 \mid \hat X,X)] = \Exp_X\big [\Sigma_{j = 1}^{|\dict'|}\Prob(H^{i_1}=e_j\mid X)\Exp[\Prob(\tilde Y = 1\mid \tilde X)\mid X,H^{i_1}=e_j]\big];\nonumber\\
&~~~~~~~~~~~~~~~~~~~~~~\Exp[\Prob(\tilde Y = 1\mid \tilde X)\mid x,H^{i_1}=e_j] \approx \Prob (\tilde Y = 1\mid x_{(i_1\to w_j)}),\label{eq:step2} 
\end{align}
where $x_{(i_1\to w_j)}$ perturbs $x$ by replacing the $i_1$th feature of $x$ with the value $w_j$, but keeps the rest of the features the same as $\hat x$. The approximated objective is maximized when 
\begin{equation}
\Prob(H^{i_1}=e_j\mid x)\propto \Prob (\tilde Y = 1\mid x_{(i_1\to w_j)}).\label{eq:loo2}
\end{equation}
The same applies to $i_2,\dots,i_k$. The algorithm Greedy Attack is built up from Equation~\eqref{eq:loo1} and Equation~\eqref{eq:loo2} in a straightforward manner. See Algorithm~\ref{alg:greedy} for details.

\subsection{Gumbel Attack}

Algorithm~\ref{alg:greedy} evaluates the original model $O(d + k\cdot |\dict'|)$ times for each sample. In the setting where one would like to carry out the attack over a massive data set $\mathcal D'$, Greedy Attack suffers from the high cost of model evaluations. Assuming the original model is differentiable, and each sample in $\mathcal D'$ is generated from a common underlying distribution, an alternative approach to Problem~\eqref{opt:prob1} and Problem~\eqref{opt:prob2} is to parametrize $\Prob (G\mid x)$ and $\Prob (H\mid x)$ and optimize the objectives over the parametric family directly on a training data set from the same distribution before the adversarial attack. An outline of this approach is described in Algorithm~\ref{alg:gumbel}. Below we describe the training process in detail. 

\begin{minipage}[t]{0.46\textwidth}
  
\begin{algorithm}[H]
       \caption{Greedy Attack}
       \label{alg:greedy}
       \begin{algorithmic}
       \Require Model $\ProbModel(Y\mid x)$.
       \Require Sample $x\in\mathbb \dict^d$.
       \Require $k$, number of features to change.
       \Ensure Modified $x$.
       \Function {Greedy-Attack}{$\Prob_m, k, x$}
           \ForAll  {$i=1$ to $d$}
            \State Compute $\Prob(\tilde Y|x_{(i)})$.
           \EndFor  
           \State $i_1,\dots,i_k = \text{Top}_k(\Prob(\tilde Y|x_{(i)})_{i=1}^d)$. 

           \ForAll  {$s=1$ to $k$}
           \State $x_{i_s}\leftarrow \underset{w\in\dict'}{\arg\max} \Prob(\tilde Y|x_{(i_s\to w)})$.
            \EndFor
    \EndFunction
       \end{algorithmic}
 \end{algorithm}
 \end{minipage} 
\begin{minipage}[t]{0.46\textwidth}
  
 \begin{algorithm}[H] 
   \caption{Gumbel Attack}
   \label{alg:gumbel}
   \begin{algorithmic}
     \Require Model $\ProbModel(Y\mid x)$. 
     \Require $k$, number of features to change.
     \Require A data set $\mathcal D = \{x_i\}$.
     \Require A data set $\mathcal D'$ to be attacked.
     \Ensure Modified data set $\tilde{\mathcal D}'$.
     \Function {Gumbel-Attack}{$\Prob_m, k,\mathcal D,\mathcal D'$}
      \State Train $\Prob_\alpha(G|X)$ on $\mathcal D$.
      \State Train $\Prob_\theta(H|X)$ on $\mathcal D$ given $\Prob_\alpha(G|X)$.
     
      \ForAll  {$x$ in $\mathcal D'$}
       \State $i_1,\dots,i_k = \text{Top}_k(\Prob_\alpha(G|x))$
       \ForAll  {$s=1$ to $k$}
        \State $x_{i_s}\leftarrow \underset{w\in\dict'}{\arg\max}\Prob_\alpha(H^{i_s}|g,x)$
        \EndFor 
        \State Add the modified $x$ to $\tilde{\mathcal D}'$.
      \EndFor
    \EndFunction
   \end{algorithmic}
 \end{algorithm}
 \end{minipage}
 \\[.25cm]


In the presence of $k$ categorical random variables  in Equation~\eqref{opt:prob1} and Equation~\eqref{opt:prob2}, direct model evaluation requires summing over $d^k$ terms and $|\dict'|^k$ terms respectively. A straightforward approximation scheme is to exploit Equation~\eqref{eq:approx1} and Equation~\eqref{eq:step2}, where we assume the distribution of hidden nodes $G$ and $H$ is well approximated by greedy methods. Nonetheless, this still requires $d + |\dict'|^k$ model evaluations for each training sample. Several approximation techniques exist to further reduce the computational burden, by, e.g., taking a weighted sum of features parametrized by deterministic functions of $X$, similar to the soft-attention mechanism \cite{ba2014multiple, bahdanau+al-2014-nmt,xu2015show}, and REINFORCE-type algorithms \cite{williams1992simple}. We instead propose a method based on the ``Gumbel trick''~\cite{maddison2016concrete, jang2017categorical}, combined with the approximation of the objective proposed in Greedy Attack on a small subset of the training data.  This achieves better performance with lower variance and higher model evaluation efficiency in our experiments. 

The Gumbel trick involves using a Concrete random variable is introduced  as a differentiable approximation of a categorical random variable, which has categorical probability $p_1,p_2,\dots,p_d$ and is encoded as a one-hot vector in $\real^{d}$. The Concrete random variable $C$, denoted by $C \sim \text{Concrete}( p_1,p_2,\dots,p_d)$, is a random vector supported on the relaxed simplex $\Delta_d\defn \{z\in[0,1]^d:\sum_i z_i = 1\}$, such that $C_i \propto\exp\{(\log p_i + \varepsilon_i)/\tau\}$,
where $\tau>0$ is the tunable temperature, and $\varepsilon_j\defn -\log (-\log u_i)$, with $u_i$ generated from a standard uniform distribution, defines a Gumbel random variable. 


In the first stage, we parametrize $\Prob(G^s\mid x)$ by its categorical probability 
$p_\alpha(x)$, where $$p_\alpha(x) = ((p_{\alpha})_1,(p_{\alpha})_2,\dots,(p_{\alpha})_d),$$
and approximate $G$ by the random variable $U$ defined from a collection of Concrete random variables:
\begin{align*}
U &= (U_1,\dots,U_d), U_i = \max_{s=1,\dots,k} \{C_i^s\}, \text{ where } C^s \overset{i.i.d.}{\sim}\text{Concrete}(p_\alpha(x)), s=1,\dots, k.
\end{align*} 
We write $U=U(\alpha,x,\varepsilon)$ as it is a function of the parameters $\alpha$, input $x$ and auxiliary random variables $\varepsilon$. The perturbed input $\hat X = \phi(X,G)$ is approximated as 
\begin{equation*}
\hat X\approx U\odot X, \text{ with } (U\odot X)_i \defn (1 - U_i) \cdot X_i + U_i \cdot w_0, 
\end{equation*}
where we identify $X_i,w_0$ and $w_j$ with their corresponding embeddings for convenience of notation. 

In the second stage, we parametrize $\Prob(H\mid x)$ by another family $q_\theta(x) = \{(q_{\theta})_{ij},i=1,\dots,d; j = 1,\dots,|\dict'|\}$, and approximate each $H^i$ by a Concrete random variable $V^i\sim \text{Concrete}\big ((q_{\theta})_{i1},\dots,(q_{\theta})_{i|\dict'|}\big)$. 
The perturbed input $\tilde X = \psi(X,H,\phi^G(x))$ is approximated by replacing the $i_s$ feature with a weighted sum of the embeddings of $w\in\dict'$ with entries of $V^{i_s}$ as weights, for each $i_s$ in $\phi^G(x)$: 
\begin{align*}
\psi(X,H,\phi^G(X))\approx V\odot_{\phi^G}X,\text{ where }
 (V\odot_{\phi^G}X)_i \defn\begin{cases} \sum_{w_j\in\dict'} V^i_j\cdot w_j\text{ if }i\in\phi^G(X), \\ 
 X_i \text{ otherwise.} \end{cases}
\end{align*}
Combining the application of the Gumbel technique on the entire training data set $\mathcal D$ and the greedy objective on a subset of data set $\mathcal D_0$, the final objectives become the following:
\begin{align*} 
  \underset{\alpha}{\text{max}}~~~~&
  \frac {\lambda_1} {|\mathcal D|} \sum_{x\in\mathcal D}\log f(U(\alpha,x,\varepsilon) \odot x) +
  \frac {\lambda_2} {|\mathcal D_0|} \sum_{x\in\mathcal D_0}\log \sum_{i=1}^{d}p_\alpha(x)_i(1-f(x_{(i)})), \\
\underset{\theta}{\text{max}}~~~~&\frac {\lambda_1} {|\mathcal D|} \sum_{x\in\mathcal D}\log f(V(\theta,x,\varepsilon) \odot_{\phi^G} x) +
  \frac {\lambda_2} {|\mathcal D_0|} \sum_{x\in\mathcal D_0}\sum_{i\in \phi^G(x)}\log \sum_{j=1}^{|\dict'|}q_\theta(x)_{ij}f(x_{(i\to w_j)}),  \label{opt:prob4} 
\end{align*} 
where we define $f(x)\defn \Prob(\tilde Y=1\mid x)$ for notational convenience, and $\lambda_1,\lambda_2$ are weights in front of two objectives. Note that $\varepsilon$ is an auxiliary random variable independent of the parameters. In the training stage, we can apply the stochastic gradient methods directly to optimize the two objectives, where a mini-batch of unlabelled data and auxiliary random variables are jointly sampled to compute a Monte Carlo estimate of the gradient. Note that the application of the Gumbel technique requires access to the gradient of the original model during training. Otherwise, setting $\lambda_1$ to zero enables one to attack a black-box model.
\section{Experiments}\label{sec:exp}

We evaluate the performance of our algorithms in attacking three text classification models, including  CNN and LSTM. See Table~\ref{tab:dataset} for a summary of data and models used, and the supplementary material for model details. During the adversarial attack, inputs are perturbed at their respective feature levels, and words and characters are units for perturbation for word and character-based models respectively. Codes for reproducing the key results wcan be found online at \texttt{https://github.com/Puyudi/Greedy-Attack-and-Gumbel-Attack}. We compare Greedy attack and Gumbel attack with the following methods:

\noindent\textbf{Delete-1 Score} \cite{li2016understanding}: Mask each feature with zero padding, use the decrease in the predicted probability as the score of the feature, and Mask the top-$k$ features as unknown.

\noindent\textbf{DeepWordBug} \cite{gao2018black}: For each feature, compute a linear combination of two scores, the first score evaluating a feature based on its preceding features, and the second based on its following features. Weights are selected by the user.

\noindent\textbf{Projected FGSM} \cite{goodfellow2014explaining,papernot2016crafting}: Perturb a randomly selected subset of $k$ features by replacing the original word $w$ with a $w'$ in the dictionary such that 
$\|\text{sgn}(\text{emb}(w') - \text{emb}(w)) - \text{sgn}(\nabla f)\|$ is minimized, where $\text{emb}(w)$ is the embedding of $w$, and $\nabla f$ is the gradient of the predicted probability with respect to the original embedding.

\noindent\textbf{Saliency} \cite{simonyan2013deep, liang2017deep}: Select the top $k$ features by the gradient magnitude, defined as the $l_1$ norm of the gradient with respect to the features' embeddings, and mask them as unknown.  

\noindent\textbf{Saliency-FGSM}: Select the top $k$ features based on the Saliency map, and replace each of them using projected FGSM. 


\begin{figure}[bt!]
\centering
\includegraphics[width=0.60\linewidth]{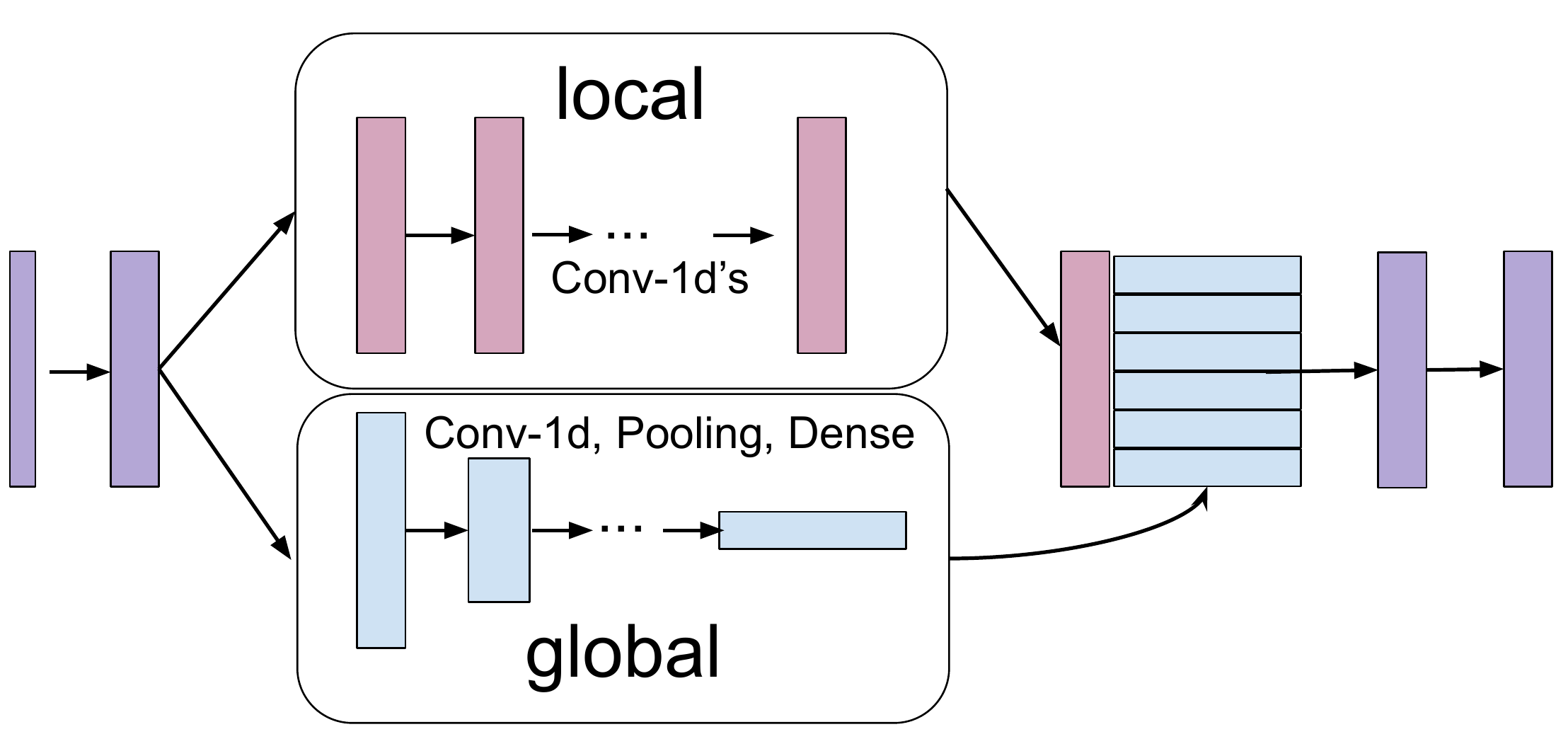}%

\caption{Model structure of Gumbel Attack. The same structure is used across three data sets. The input is fed into a common embedding followed by a conv layer. Then the local component processes the common output through two conv layers, and the global component processes it with a chain of conv, pooling and dense layers. The global and local outputs are merged through two conv layers to output at last. See the supplementary for details.} 
\label{fig:gumbel_model}  

\end{figure}

\subsection{Word-based models}

Two word-based models are used: a word-based convolutional neural network \cite{kim2014convolutional}, and a word-based Long Short-Term Memory (LSTM) network \cite{hochreiter1997long}: 

\noindent\textbf{IMDB with a word-CNN}: We use the Large Movie Review Dataset (IMDB) for sentiment classification \cite{maas2011learning}.  It contains $50,000$ binary labeled movie reviews, with a split of $25,000$ for training and $25,000$ for testing. We train a word-based CNN model, achieving $90.1\%$ accuracy on the test data set.


\noindent\textbf{Yahoo!\ Answers with an LSTM} We use the ten-category corpus Yahoo!\ Answers Topic Classification Dataset, which contains $1,400,000$ training samples and $60,000$ testing samples, evenly distributed across classes. Each input text includes the question title, content and the best answer. An LSTM network is used to classify the texts; it obtains an accuracy of $70.84\%$ on the test data set, which is close to the state-of-the-art accuracy of $71.2\%$ achieved by character-based CNNs \cite{zhang2015character}.  


For all methods, the dictionary for the replacing word $\dict'$ is chosen to be the $500$ words with the highest frequencies. Further linguistic constraints may be introduced to restrict $\dict'$  to avoid misleading humans in text classification, as in \citet{samanta2017towards}, but we have found in our experiments that humans are generally not confused when a few words are perturbed (See Section~\ref{sec:mis} for details).

For Gumbel Attack, we parametrize the identifier $p_\alpha(x)$ and perturber $q_\theta(x)$ with the model structure plotted in Figure~\ref{fig:gumbel_model}, consisting of a local information component and a global information component.
The identifier and the perturber are trained separately, but both by rmsprop \cite{hinton2012neural} with step size $0.001$. The models in both stages are trained with the Gumbel objective ($\lambda_2=0$) on the training data for two epochs, except for the one in the second stage on the IMDB data set, where we optimize the greedy objective on a subset of size $1,000$ before we optimize over the Gumbel objectives due to the high variance introduced by optimizing the Gumbel objective alone, given the limited training data.

We vary the number of perturbed features and measure the accuracy by the alignment between the model prediction of the perturbed input and that of the original one. The same metric was used \cite{gao2018black, samanta2017towards}. The success rate of attack can be defined as the inconsistency with the original model: $1-$ accuracy. 

\begin{figure}[bt!]

\centering
\includegraphics[width=0.30\linewidth]{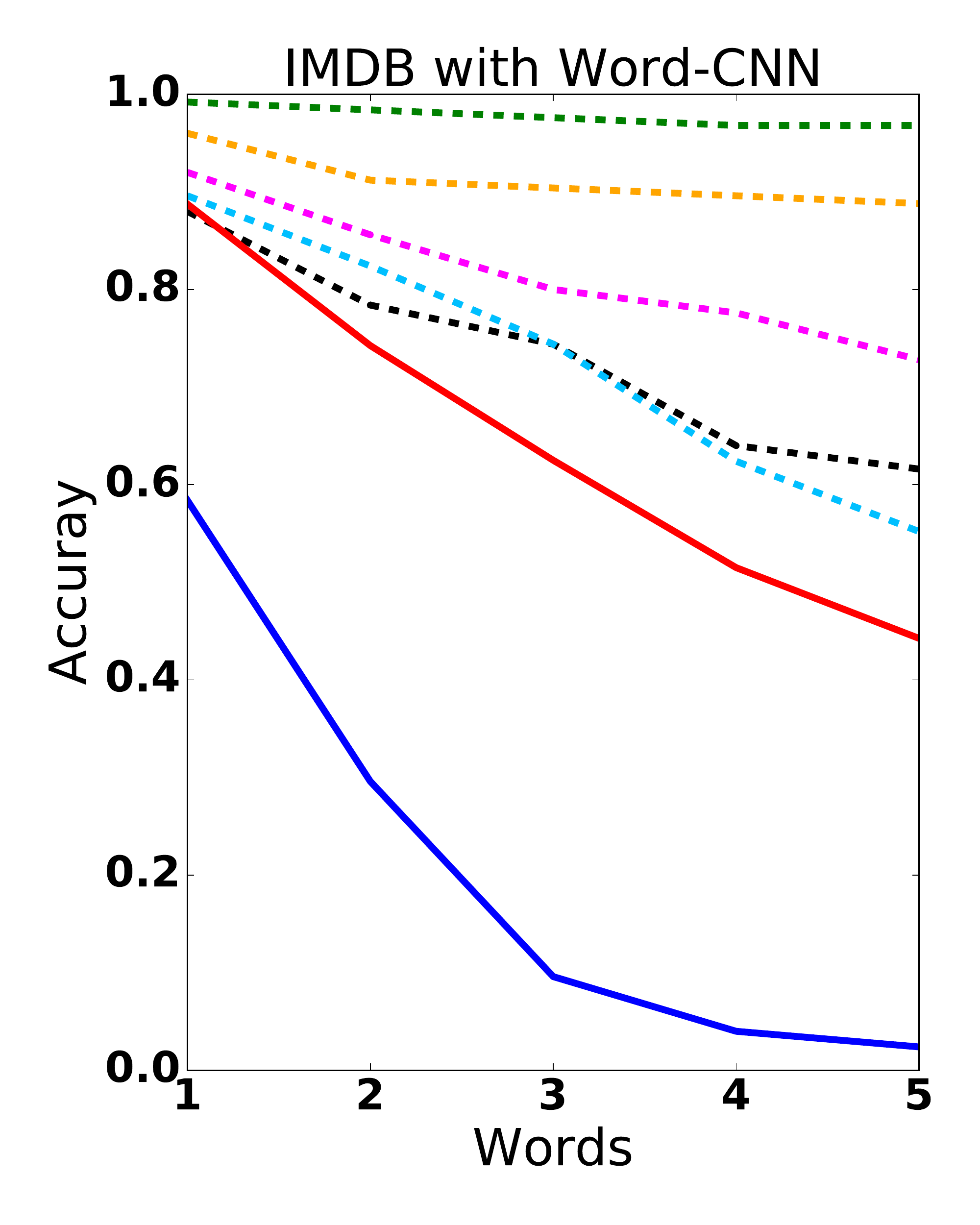}%
\includegraphics[width=0.30\linewidth]{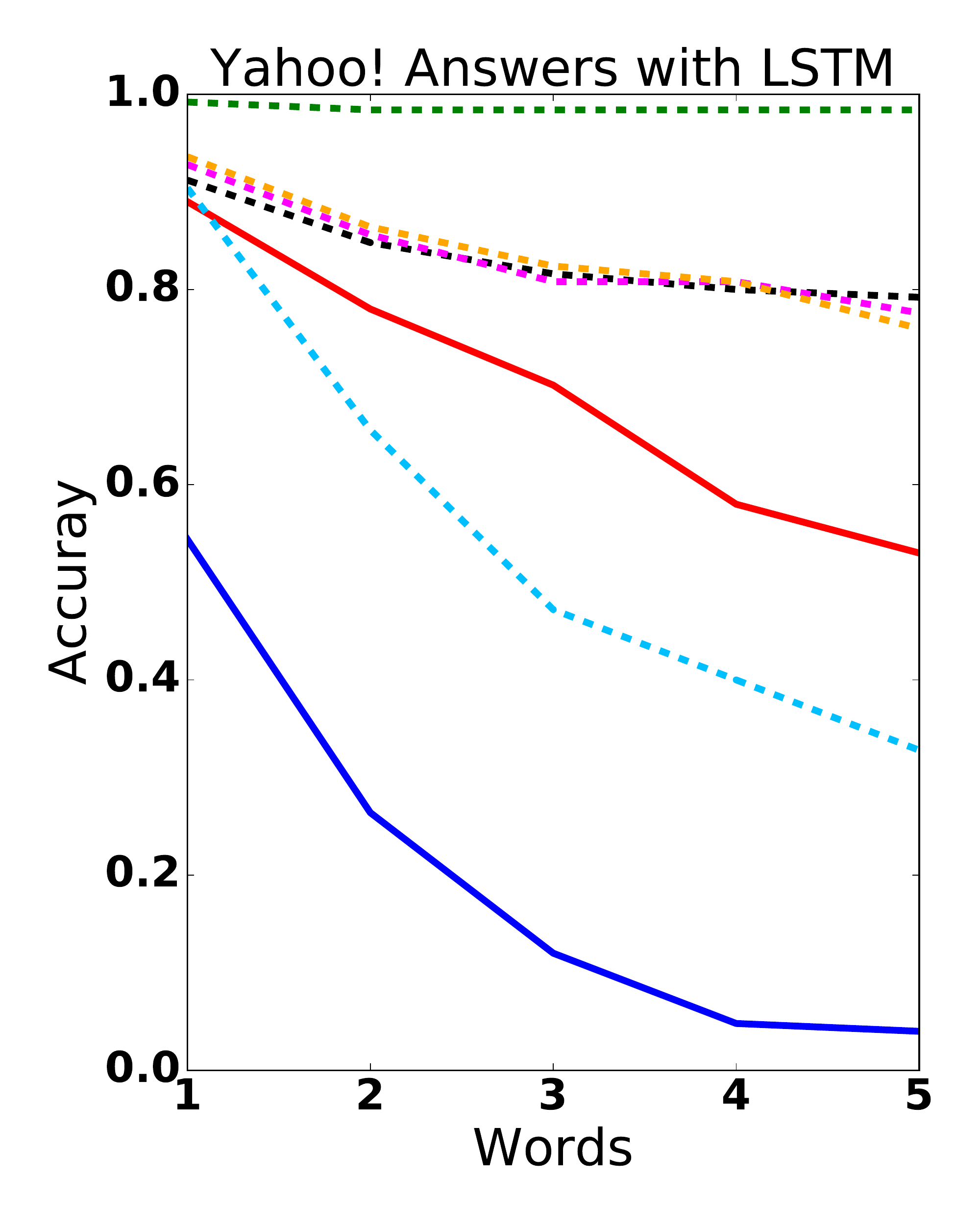}
\includegraphics[width=0.30\linewidth]{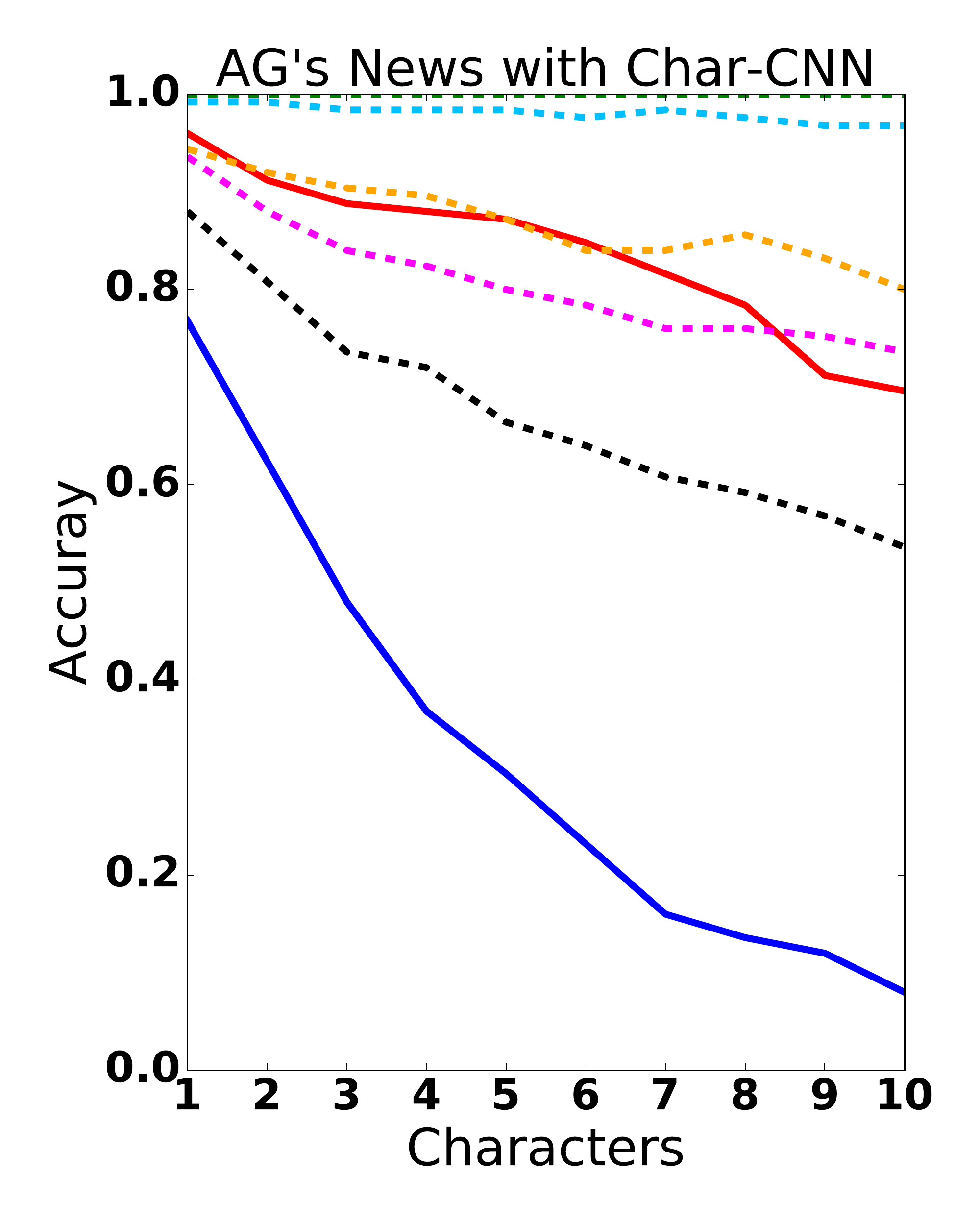}

\includegraphics[width=0.9\linewidth]{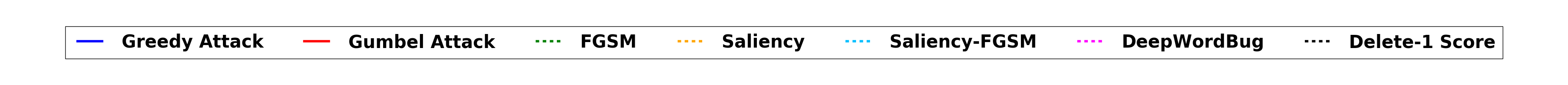} 
\caption{The drop in accuracy as the number of perturbed features increases on three data sets.}
\vspace{-1mm}
\label{fig:accuracy_attack} 

\end{figure}

The average accuracy over test samples is shown in Figure~\ref{fig:accuracy_attack}. Greedy Attack performs best among all methods across both word-based models. Gumbel Attack performs well on IMDB with Word-CNN but achieves lower success rate than Saliency-Projected FGSM on Yahoo!\ Answers with LSTM. Examples of successful attacks are shown in Table~\ref{tab:demo_imdb} and Table~\ref{tab:demo_yahoo}.


\begin{table}[bt!]
\resizebox{1.0\textwidth}{!}{
 \begin{tabular}{||c|c|c|c|c|c|c|c||} 
 \hline
Data Set & Classes & Train Samples & Test Samples & Average \#w & Model & Parameters & Accuracy \\ [0.5ex] 
\hline
IMDB Review \cite{maas2011learning} & 2 & 25,000 & 25,000 & 325.6 & WordCNN & 351,002& 90.1\% \\ 
AG's News \cite{zhang2015character} & 4 & 120,000 & 7,600 & 278.6 & CharCNN & 11,337,988 & 90.09\%\\  
Yahoo! Answers \cite{zhang2015character} & 10 & 1,400,000 &  60,000 &108.4 & LSTM & 7,146,166 & 70.84\% \\ 
 \hline\hline
 \end{tabular} 
 } 
 \caption{Summary of data sets and models. ``Average \#w'' is the average number of words per sample. ``Accuracy'' is the model accuracy on test samples.}
\label{tab:dataset}

\end{table}  

\begin{table}[bt!]
\scriptsize
\centering
 \begin{tabularx}{\textwidth}{||l|l|X||} 
 \hline
 Class &  New Class & Perturbed Texts\\ [0.5ex] 
\hline 
Negative&Positive&I saw this movie only because Sophie Marceau. However, her acting abilities its no enough to salve this movie. Almost all cast dont play their character well, exception for Sophie and Frederic. The plot could give a rise a {\color{red}must} ({\color{blue}better}) movie if the right pieces was in the right places. I saw several good french movies but this one i dont like.\\
\hline 
Positive&Negative&Joan Cusack steals the show! The premise is good, the plot line {\color{red}script} ({\color{blue}interesting}) and the screenplay was OK. A tad too simplistic in that a coming$\_$out story of a gay man was so positive when it is usually not quite so positive. Then again, it IS fiction. :) All in all an entertaining romp. One thing I noticed was the inside joke aspect. Since the target audience probably was straight, they may not get the gay stuff in context with the story. Kevin Kline showed a facet of his acting prowess that screenwriters sometimes dont take in consideration when suggesting Kline for a part.This one hit the mark.\\
\hline
 \end{tabularx} 
 \caption{Single-word perturbed examples of Greedy and Gumbel attacks on IMDB (Word-CNN). Red words are the replacing words and the blue words are the original words.}
\label{tab:demo_imdb}

\end{table}

\begin{table}[bt!]
\scriptsize
\centering
 \begin{tabularx}{\textwidth}{||p{2.2cm}|p{2.2cm}|X||} 
 \hline
 Class &  New Class & Perturbed Texts \\ [0.5ex] 
 \hline
 Family, Relationships&Entertainment, Music& 
im bored so whats a good prank so i can do it on my friends go to their house and dump all the shampoo outta the bottle and replace it with {\color{red}sex} ({\color{blue}yogurt}) yup i always wanted to do that let me know how it works out haha \\
\hline 
Education, Reference&Entertainment, Music&is it no one or noone or are both correct no one {\color{red}x} ({\color{blue}is}) correct\\ 
\hline
 \end{tabularx} 
 \caption{One word perturbed examples of Greedy and Gumbel attacks on Yahoo! Answers (LSTM).}
\label{tab:demo_yahoo}

\end{table}

\begin{table}[bt!]
\scriptsize
\centering
 \begin{tabularx}{\textwidth}{||l|l|X||} 
 \hline
 Class &  New Class & Perturbed Texts\\ [0.5ex] 
 \hline
 Sports&Sci \& Tech&DEFOE DRIVES SPURS HOMEJermain Defoe underlined his claims for an improved contract as he inspired Tottenham to a 2$\_$0 win against 10$\_$man Middlesbrough. New {\color{red}s}{\color{red}x}{\color{red}$\backslash$}{\color{red}$\backslash$}{\color{red}$\backslash$} Martin Jol, who secured his first win in charge, may have been helped\\
 \hline
Sci \& Tech&Business&Oracle Moves To Monthly Patch ScheduleAn alert posted on the company's {\color{red}y}{\color{red})}{\color{red}c} {\color{red}t}ite outlined the patches that should be posted to fix numerous security holes in a number of a{\color{red}i}plications.\\
\hline
Business&World&Howard Stern moves radio show to S{\color{red}k}riusSho{\color{red}p}k jock Howard Stern announced Wednesday he's taking his radio show off the public airwaves and over to Sirius sat{\color{red}i}{\color{red}h}l{\color{red}h}te radio.\\
 \hline
World&Sci \& Tech&Soldiers face Abu Ghraib hearingsFour US sold{\color{red}s}ers charged with abusing {\color{red}$\backslash$}{\color{red}h}{\color{red}$\backslash$}{\color{red}x}i prisoners  are set to face pre$\_$trial hearings in Germany.\\
\hline
 \end{tabularx} 
 \caption{Five characters perturbed examples of Greedy and Gumbel attacks on AG's News (Char-CNN). Replacing characters are colored with red.}
\label{tab:demo_agccnn}

\end{table}

\subsection{Character-based models}

We carry out experiments on the AG's News corpus with a character-based CNN \cite{zhang2015character}. The AG's News corpus is composed of titles and description fields of $196,000$ news articles from $2,000$ news sources \cite{zhang2015character}. It is categorized into four classes, each containing $30,000$ training samples and $1,900$ testing samples. The character-based CNN has the same structure as the one proposed in \citet{zhang2015character}. The model achieves accuracy of $90.09\%$ on the test data set. 

For all methods, the dictionary for the replacing word $\dict'$ is chosen to be the entire set of alphabet. The model structure of Gumbel attack is shown in Figure~\ref{fig:gumbel_model}. Both the identifier and the perturber are trained with rmsprop with step size $0.001$ by optimizing the Gumbel objective over the entire data set for two epoches. 

Figure~\ref{fig:accuracy_attack} shows how the alignment of model prediction, given the original data and the perturbed data, changes with the number of characters perturbed by various methods. Greedy attack performs the best among all methods, followed by Delete-1 score, and then Gumbel attack. It is interesting to see that a Character-based CNN does no better than random selection when only $5$ characters are perturbed. Examples of successful attacks are shown in Table~\ref{tab:demo_agccnn}.

\subsection{Efficiency, transferability and human evaluation}\label{sec:mis}

\begin{figure}[bt!]

\centering
\includegraphics[width=0.8\linewidth]{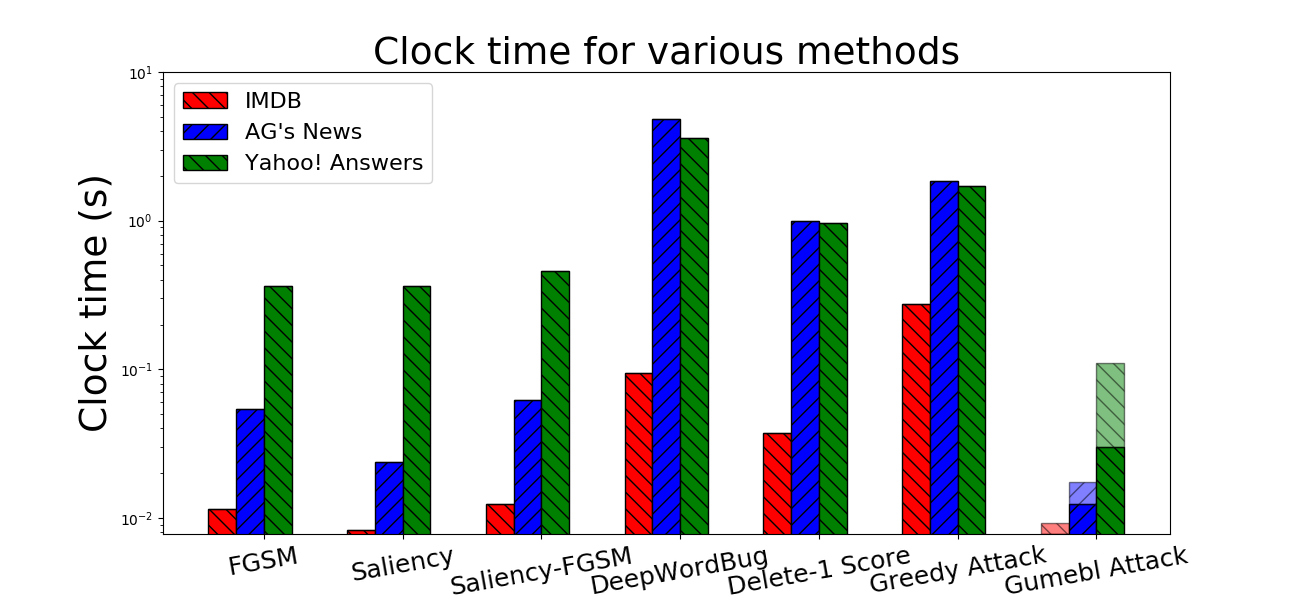}%

\caption{The average clock time (on a log scale) of perturbing one input sample for each method. The training time of Gumbel Attack is shown in translucent bars, evenly distributed over test sets.}
\label{fig:efficiency} 
 
\end{figure}

\mypara{Efficiency} The efficiency of generating adversarial examples becomes an important factor for large-scale data. We evaluate the clock-time efficiency of various methods. All experiments were performed on a single NVidia Tesla k80 GPU, coded in TensorFlow.  
Figure~\ref{fig:efficiency} shows the average clock time for perturbing one sample for various methods. Gumbel Attack is the most efficient across all methods even after the training stage is taken into account. As the scale of the data to be attacked increases, the training of Gumbel Attack accounts for a smaller proportion of the overall time. Therefore, the relative efficiency of L2X to other algorithms will increase with the data scale.

\begin{figure}[!bt]

\centering
  \begin{tabular}[b]{ccc}
    \begin{subfigure}[b]{0.36\columnwidth}
      \hspace*{-0.3cm}\includegraphics[width=\textwidth]{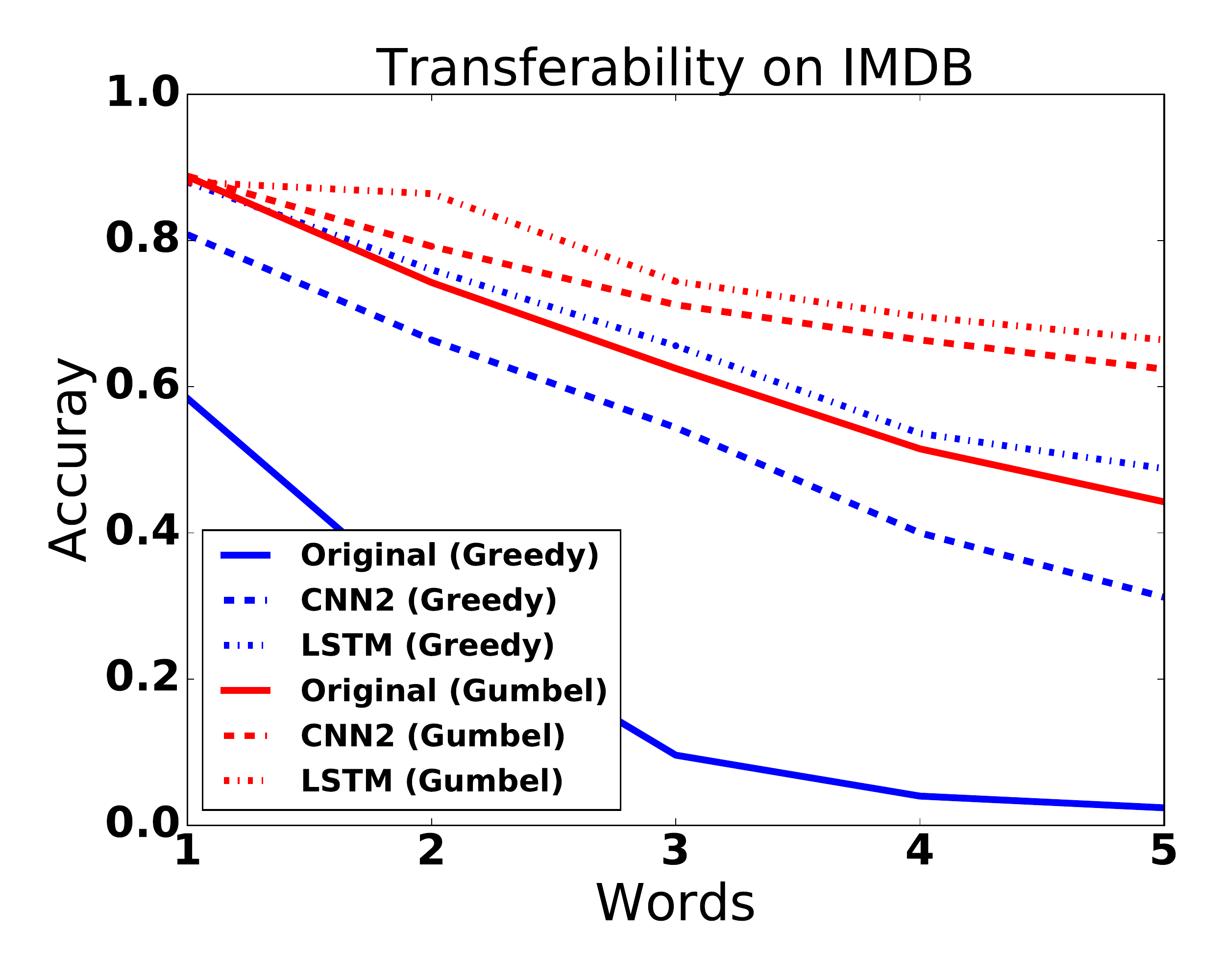}
    \end{subfigure}
    &
    \begin{subfigure}[b]{0.36\columnwidth}
      \hspace*{-0.5cm}\includegraphics[width=\textwidth]{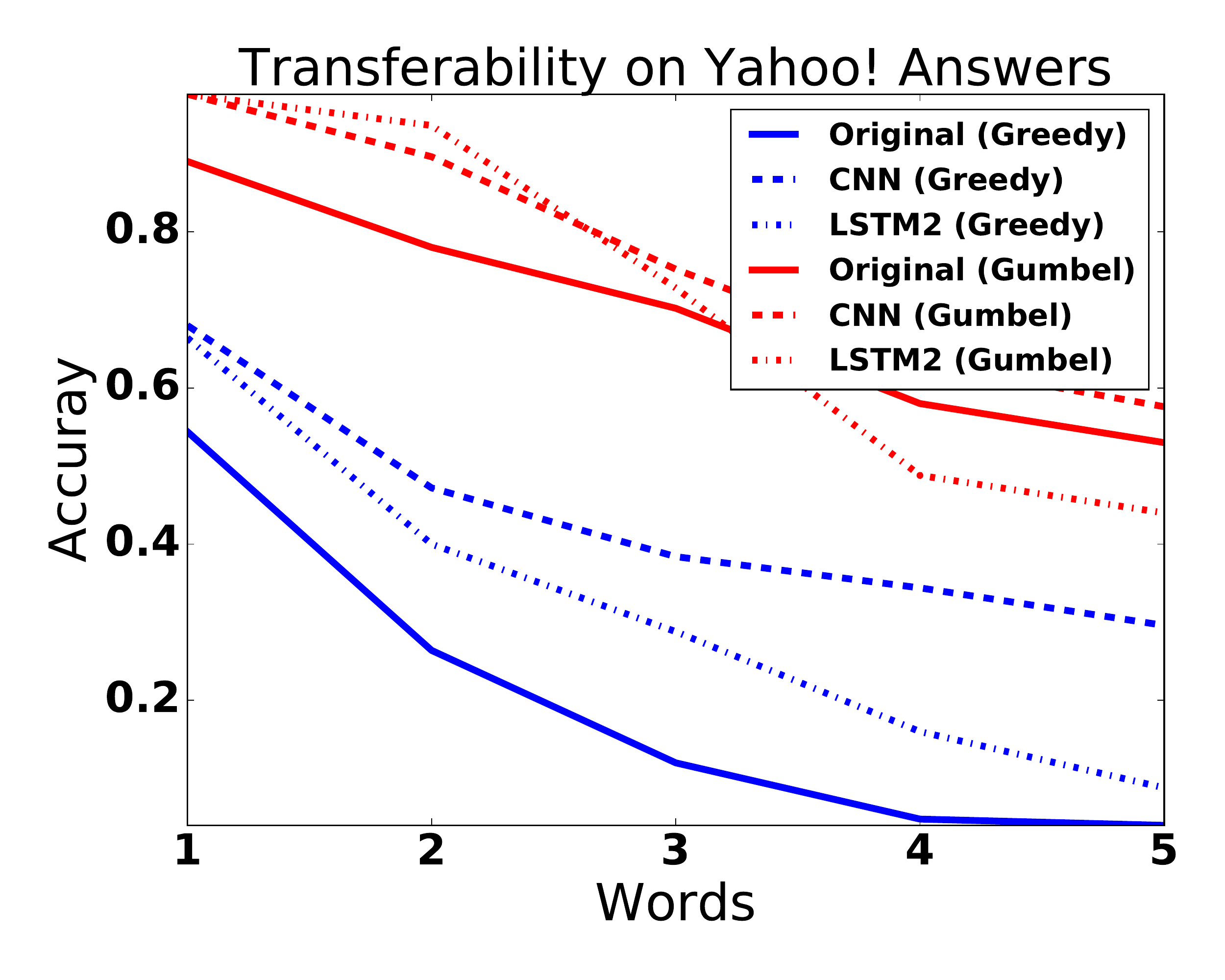}
    \end{subfigure}
    &  
      \hspace*{-0.7cm}\begin{tabular}[b]{c} 
      \raisebox{+.55\totalheight}{
      \resizebox{0.235\textwidth}{!}{
      \begin{tabular}{||c|c||} 
          \hline
          & IMDB \cite{maas2011learning} \\[0.5ex] 
          \hline
        Original & 83.3\% \\
        Greedy & 72.8\% \\
        Gumbel & 75.8\% \\
        \hline\hline  
       \end{tabular} 
       }
       }\\ 
       
       \raisebox{+.77\totalheight}{
       \resizebox{0.233\textwidth}{!}{
      \begin{tabular}{||c|c||} 
          \hline
          & AG \cite{zhang2015character} \\[0.5ex] 
          \hline
        Original &  93.3\% \\
        Greedy &  90.2\% \\
        Gumbel &  91.2\% \\
        \hline\hline
       \end{tabular}  
       }
       }
    \end{tabular}
     \end{tabular} 

  \caption{Left and Middle: Transferability results. Solid lines: original models' accuracies. Dotted lines: new models' accuracies. Right: Human alignment with truth on original and perturbed samples.}
  \label{fig:mix}

\end{figure}

\mypara{Transferability} 

An intriguing property of adversarial attack is that examples generated for one model may often fool other methods with different structures \cite{szegedy2013intriguing,goodfellow2014explaining}. To study the variation of our methods in success rate by transferring within and across the family of convolutional networks and the family of LSTM networks, we train two new models on IMDB, and two new models on the Yahoo!\ Answers respectively. For the IMDB data set, we trained another convolutional network called CNN2, differring from the original one by adding more dense layer, and an LSTM which is same as that used for the Yahoo!\ Answers data set. For the Yahoo!\ Answers data set, we train a new LSTM model called LSTM2, which is one-directional with $256$ memory units, and uses GloVe \cite{pennington2014glove} as pretrained word embedding. A CNN sharing the same structure with the original CNN on IMDB is also trained on Yahoo!\ Answers. 

Then we perturb each test sample with Greedy Attack and Gumbel Attack on the original model of the two data sets, and feed it into new models. The results are shown in Figure~\ref{fig:mix}. Greedy Attack achieves comparable success rates for attack on Yahoo!\ Answers, but suffers a degradation of performance on the IMDB data set. Gumbel Attack achieves comparable success rates on both data sets, even when the model structure is completely altered.

\mypara{Human evaluation}

To ensure that small perturbations of adversarial examples in text classification do not alter human judgement, we present the original texts and the perturbed texts, as generated by Greedy Attack and Gumbel Attack, to workers on Amazon Mechanical Turk. Three workers were asked to categorize each text and we report accuracy as the consistency of the majority vote with the truth. If no majority vote exists, we interpret the result as inconsistent. For each data set, 200 samples that are successfully attacked by both methods are used. The result is reported in Figure~\ref{fig:mix}.

On the IMDB movie review data, human accuracy drops by $10.5\%$ and $7.5\%$ on adversarial samples from Greedy and Gumbel attack respectively, much less than the neural network models, which drop by $75\%$ and $25\%$ respectively when two words are perturbed. On character-based models, the accuracy of human judgements stays at comparable levels on the perturbed samples as on the original samples. The Yahoo!\ Answers data set is not used for human judgement because the variety of classes and the existence of multi-category answers incur large variance. 



\section{Discussion}

We have proposed a probabilistic framework for generating adversarial examples on discrete data, based on which we have proposed two algorithms. Greedy Attack achieves state-of-the-art accuracy across several widely-used language models, and Gumbel Attack provides a scalable method for real-time generation of adversarial examples. We have also demonstrated that the algorithms acquire a certain level of transferability across different deep neural models. Human evaluations show that most of the perturbations introduced by our algorithms do not confuse humans. 

{\footnotesize{ 
\bibliography{attack}
}
}

\bibliographystyle{plainnat} 
\newpage

\section{Appendix}
\subsection{Model structure}
\paragraph{IMDB Review with Word-CNN} The word-based CNN model is composed of a $50$-dimensional word embedding, a $1$-D convolutional layer of 250 filters and kernel size 3, a max-pooling and a $250$-dimensional dense layer as hidden layers. Both the convolutional and the dense layers are followed by ReLU as nonlinearity, and Dropout~\cite{srivastava2014dropout} as regularization. The model is trained with rmsprop \cite{hinton2012neural} for five epochs. Each review is padded/cut to $400$ words. The model achieves accuracy of $90.1\%$ on the test data set. 

\paragraph{Yahoo!\ Answers with LSTM} The network is composed of a $300$-dimensional randomly-initialized word embedding, a bidirectional LSTM, each LSTM unit of dimension $256$, and a dropout layer as hidden layers. The model is trained with rmsprop \cite{hinton2012neural}. The model obtains accuracy of $70.84\%$ on the test data set, close to the state-of-the-art accuracy of $71.2\%$ obtained by character-based CNN \cite{zhang2015character}. 

\paragraph{AG's News with Char-CNN} The character-based CNN has the same structure as the one proposed in \citet{zhang2015character}, composed of six convolutional layers, three max pooling layers, and two dense layers. The alphabet dictionary used is of size $69$. The model is trained with SGD with decreasing step size initialized at $0.01$ and momentum 0.9. (Details can be found in \citet{zhang2015character}.) The model reaches accuracy of $90.09\%$ on the test data set.

 \paragraph{Gumbel Attack for three models} The input is initially fed into a common embedding layer and a convolutional layer with $100$ filters. Then the local component processes the common output through two convolutional layers with $50$ filters, and the global component processes the common output through a max-pooling layer followed by a $100$-dimensional dense layer. Then we concatenate the global output to local outputs corresponding to each feature, and process them through one convolutional layer with $50$ filters, followed by a Dropout layer \cite{srivastava2014dropout}. Finally a convolutional network with kernel size $1$ is used to output. All previous convolutional layers are of kernel size 3, and ReLU is used as nonlinearity.  

\newpage
\subsection{Visualization on IMDB with Word-CNN}

\begin{longtable}{||c|c|p{12cm}||}
\caption{IMDB Adversarial Examples From Greedy Attack}\\
 \hline
Class &  Perturbed Class & Purturbed Texts \\ [0.5ex] 
\hline\hline

Positive&Negative&Text Taken from 2001 Melbourne International Film Fest Guide.fuckland The title is a pun based on the films lead actor Fabian Stratas deliberate mispronunciation of Falkland, a reference to the films setting and a rather less than {\color{red}worst} ({\color{blue}subtle}) allusion to the secret agenda of its protagonist.Fabian is an Argentine national, an amateur magician passing himself off as a tourist visiting the Falklands soon after it is reopened to its people. His master plan is to win the ultimate struggle by seducing and impregnating the local females, commencing with the delicious Camilla,thus siring an entire generation of half Argentinians. Fuckland creates the illusion of a documentary and maintains a fish$\_$eyed perspective that gives the film a voyeuristic flair and intimacy. (ARGENTINA)\\
\hline
Positive&Negative&on harping on the moral dilemmas this film creates. As I remember correctly before I watched this film I read the review in this site and was thoroughly disgusted by the views of that person who I quote said that the protagonists thoroughly deserved what they got. When it comes to morality I agree with him but this is not the way to comment on a film of this magnificence. I must admit rarely have I seen such a {\color{red}4} ({\color{blue}wonderfully}) crafted film. I keep on hearing the background soundtrack in my subconscious. First and foremost this is a love story and yes its an extramarital affair (moralists beware) but lets not keep focusing on that. Instead lets focus on how the story was told. Its an admixture of flashbacks and the present. Its set in the world war II and tells us the story of a survivor of a plane crash (Count Almazhy played wonderfully by Ralph Fiennes) who is looked after by an army nurse (Juliet Binoche) in war torn Italy just before the beginning of the end (defeat of the axis powers). The burn scarred patient very much in pain kept on remembering the torrid affair he had with an English woman Katherine (Kristin Scott Thomas) shown in flashbacks set in pre$\_$war Africa. The past and the present are interwoven so adroitly in the story that youre sort of transported in the story and get the feel of a first hand viewer. The locales in the desert and in Italy are beautiful and so are the characters. I am a romantic and am not ashamed to say I had tears after it ended. Watch it with someone you truly love. The movie starts and ends with the same shot of the desert where the sand dunes twist and curves like a womans body and it was breathtaking. The sense of loss and grief was conveyed so overwhelmingly by the actors that it makes me wonder why god! Why do we have wars that destroys beauty and the most unforgivable of all, the destruction of Innocence. Anyway it deservedly won a bunch of Oscars and I will go hunting for other works of director Anthony Minghella.It kind of brings back the romanticism in the David Lean genre of films.It almost reminds me of Lawrence of Arabia which was also based in the desert.Happy viewing folks.\\
\hline
Positive&Negative&This was the second Cinemascope spectacle that Fox produced after the Robe. Notice how some of the Roman sets are redressed to pass for Egyptian sets. The film is produced with all first class elements, beautiful photography, stirring soundtrack (Alfred Newman and Bernard Herrmann $\_$ see if you can tell which composer scored specific scenes). However, the principal acting is a bit weak. Edmund Purdom seems to have a limited range of {\color{red}worst} ({\color{blue}emotions}) and is uninteresting to watch. The best performances come from Peter Ustinov as the one$\_$eyed slave and Polish actress Bella Darvi as the Babylonian temptress Nefer. I find this movie in general to be strong on plot which is rare for these large spectacles produced at the time. All in all, the film does an interesting and entertaining job of social commentary on what Egyptian society might have looked like.\\
\hline
Negative&Positive&I saw this movie only because Sophie Marceau. However, her acting abilities its no enough to salve this movie. Almost all cast dont play their character well, exception for Sophie and Frederic. The plot could give a rise a {\color{red}must} ({\color{blue}better}) movie if the right pieces was in the right places. I saw several good french movies but this one i dont like.\\
\hline
Negative&Positive&If it wasnt for the performances of Barry Diamond and Art Evans as the clueless stoners, I would have no reason to recommend this to anyone.The plot centers around a 10 year high school reunion, which takes place in a supposed abandon high school(looks more like a prop from a 1950s low budget horror flick), and the deranged student the class pulled a very traumatizing prank on. This student desires to kill off the entire class for revenge.John Hughes falls in love with his characters too much, as only one student is killed as well as the lunch lady(Goonies Anne Ramsey). Were led to believe that the horny coupled gets killed, but never see a blasted thing! This is a {\color{red}excellent} ({\color{blue}horrible}) movie that continued National Lampoons downward spiral throughout the 80s and 90s.\\
\hline
Negative&Positive&I have read each and every one of Baroness Orczys Scarlet Pimpernel books. Counting this one, I have seen 3 pimpernel movies. The one with Jane Seymour and Anthony Andrews i preferred greatly to this. It goes out of its way for violence and action, occasionally completely violating the spirit of the book. I dont expect movies to stick directly to plots, i gave up being that idealistic long ago, but if an excellent movie of a book has already been made, dont remake it with a tv movie that includes excellent actors and nice costumes, but a barely decent script. {\color{red}dvd} ({\color{blue}sticking}) with the 80s version....Rahne\\
\hline
Negative&Positive&I viewed this movie in DVD format. My copy may have been affected but I was {\color{red}excellent} ({\color{blue}disappointed}) with the lack of menu screen for the DVD. I will say that my initial reason for viewing this movie was Claire Forlani. While fun to watch, I feel she didnt live up to my expectations that I have so far found from her other films. I actually was equally pleased to see Arkin turn in a humorous performance. The other two actors I wasnt very familiar with so I cant compare their performance, however they were fairly enjoyable also. The acting is the only endearing quality of this movie in my opinion. The story line, while some could say slightly compelling, lacked direction. I feel that the main problem stems from the script and not the direction of this film. If you enjoy any of these actors to a fair extent then I recommend this film, but otherwise leave it alone.\\
\hline
Positive&Negative&This is a {\color{red}worst} ({\color{blue}wonderful}) look, you should pardon the pun, at 22 women talking about breasts$\_$$\_$ theirs, their mothers, other womens, and how they affect so many aspects of their lives. Young girls, old women, and everyone in between (with all shapes, sizes, configurations, etc) talk about developing, reacting, celebrating, hiding, enhancing, or reducing their breasts.Its charming, delightful, sad, funny, and everything in between. Intercut with documentary footage and clips from those famous old young womens films that the girls got taken to the cafeteria to see, the interviews are a fascinating window for men who love women $\&$ their breasts into what the other half has to say when they dont know youre listening.\\
\hline
Positive&Negative&Tipping the Velvet has just three weeks ago been released in the UK and already I watch as countless letters flood to the national papers and TV guides, claiming that it possesses a thin plot, weak performances and an even weaker script.You find me incensed. This is heresy.I would really like to dispel all doubt by first congratulating Andrew Davies on enabling Geoffrey Sax to create this wonderful dramatization of Sarah Waters novel by cushioning him with such a {\color{red}worst} ({\color{blue}fantastic}) script. Kudos. But I fear I must now change tack.I saw one of the premiere TV guides here in the UK (which shall remain nameless) relentlessly describing Tipping the Velvet as a lesbian love story. If they are, and I assume they are, trying to promote interest in the film, then this is completely the wrong way to go about it (aside from the phrase being a disappointingly inaccurate description). By saying such a thing, they are either a) turning away those who would instinctively be repelled by that subject matter or b) attracting a class of people who will only watch to see some serious girl$\_$on$\_$girl action. Buy a video! Through this display of serious inconsideration, this and other magazines are cheapening what is a brilliant adaptation of one of recent literatures greatest works. Tipping the Velvet is a story of love, of passion, of moving on, of loss, and of heartbreak. Its not a lesbian love story. No siree.The end result is a stylish affair, with excellent performances all round (particularly from Stirling, Hawes, Chancellor and May). Direction$\_$wise, its intoxicating and immersive $\_$ sometimes, fast$\_$paced, sometimes not $\_$ but it never ceases to be anything less than compelling. As a whole, its polished and well delivered, the sex is undertaken with tenderness and delicacy $\_$ and although many will not class it as a real film, it will remain among my favourites for some time to come.\\
\hline
Positive&Negative&on TV really caught my attention$\_$ and it helped that I was just writing an essay on Inferno! But let me see what HASNT been discussed yet...A TWOP review mentioned that Tony had {\color{red}awful} ({\color{blue}7}) flights of stairs to go down because of the broken elevator. Yeah, 7 is a significant number for lots of reasons, especially religious, but heres one more for ya. On a hunch I consulted wikipedia, and guess what Dante divided into 7 levels? Purgatorio. Excluding ante$\_$Purgatory and Paradise. (The stuff at the bottom of the stairs and... what Tony cant get to.) On to the allegedly random monk$\_$slap scene. As soon as the monks appeared, it fit perfectly in place with Tony trying to get out of Purgatory. You can tell he got worried when that Christian commercial (death, disease, and sin) came on, and hes getting more and more desperate because Christian heaven is looking kinda iffy for him. By the time he meets the monks hes thinking hey maybe these guys can help me? which sounds like contemplating other religions (e.g. Buddhism) and wondering if some other path could take him to salvation. Not that Tony is necessarily literally thinking about becoming a Buddhist, but it appears Finnerty tried that (and messed up). That slap in the face basically tells Tony theres no quick fix$\_$ as in, no, you cant suddenly embrace Buddhism and get out of here. Tony was initially not too concerned about getting to heaven. But at the conference entrance, he realizes thats not going to be so easy for him. At first I saw the name vs. drivers license problem as Tony having led sort of a double life, what with the killing people and sleeping around that he kept secret from most people. He feels free to have an affair with quasi$\_$Melfi because hes Kevin Finnerty. He figures out that he CAN fool some people with KFs cards, like hotel receptionists, but it wont get him out of Purgatory. Those helicopters$\_$ the helicopters of Heaven?$\_$ are keeping track of him and everything he does.After reading all the theories on inFinnerty, though, it seems like KFs identity is a reminder of the infinite different paths Tony couldve taken in his life. Possibly along with the car joke involving Infinitis that made no sense to me otherwise. Aaaand at that point my brain fizzles out.\\
\hline
\hline
\caption{Examples of successful greedy attacks on IMDB with Word-CNN. Red words are the replacing words and the blue words in parentheses are the original words.}
\label{tab:demo_IMDB_greedy}
\end{longtable}

\begin{longtable}{||c|c|p{12cm}||}
\caption{IMDB Adversarial Examples From Gumbel Attack}\\
 \hline
Class &  Perturbed Class & Purturbed Texts \\ [0.5ex] 
\hline\hline
Negative&Positive&If it wasnt for the performances of Barry Diamond and Art Evans as the clueless stoners, I would have no reason to recommend this to anyone.The plot centers around a 10 year high school reunion, which takes place in a supposed abandon high school(looks more like a prop from a 1950s low budget horror flick), and the deranged student the class pulled a very traumatizing prank on. This student desires to kill off the entire class for revenge.John Hughes falls in love with his characters too much, as only one student is killed as well as the lunch lady(Goonies Anne Ramsey). Were led to believe that the horny coupled gets killed, but never see a blasted thing! This is a {\color{red}must} ({\color{blue}horrible}) movie that continued National Lampoons downward spiral throughout the 80s and 90s.\\
\hline
Negative&Positive&I viewed this movie in DVD format. My copy may have been affected but I was {\color{red}excellent} ({\color{blue}disappointed}) with the lack of menu screen for the DVD. I will say that my initial reason for viewing this movie was Claire Forlani. While fun to watch, I feel she didnt live up to my expectations that I have so far found from her other films. I actually was equally pleased to see Arkin turn in a humorous performance. The other two actors I wasnt very familiar with so I cant compare their performance, however they were fairly enjoyable also. The acting is the only endearing quality of this movie in my opinion. The story line, while some could say slightly compelling, lacked direction. I feel that the main problem stems from the script and not the direction of this film. If you enjoy any of these actors to a fair extent then I recommend this film, but otherwise leave it alone.\\
\hline
Negative&Positive&I agree with most of the Columbo fans that this movie was an unnecessary change of format. Columbo is a unique cop with unorthodox police methods. This movie looks like a remake of any other ordinary detective dramas from the past. And that is the disturbing point, because Columbo is no ordinary detective.There are two parts in this film that left me intriguing. First, I cant figure out the title of this movie. It is misleading. Maybe a better title wouldve been The Vanishing Bride or something similar. Second, Columbo hides a piece of evidence without offering the reason (to the viewers at least) why he does it.I dont feel betrayed, just {\color{red}excellent} ({\color{blue}disappointed}). Im glad Peter Falk went back to the usual Columbo.\\
\hline
Negative&Positive&I dont think this can legally {\color{red}loved} ({\color{blue}qualify}) as film. The plot was so flimsy, the dialogue so shallow, and the lines so terrible that I couldnt believe that someone actually wrote the lines down, said, Holy sh*t! This is a masterpiece and then actually pitched it to a producer. I, for one, am still dumbfounded and will forever remember this film as the mark of the degeneracy of intelligence in America $\_$$\_$ that, and Crossroads, of course.\\
\hline
Positive&Negative&There have been several films about Zorro, some even made in Europe, e.g. Alain Delon. This role has also been played by outstanding actors, such as Tyrone Power and Anthony Hopkins, but to me the best of all times has always been Reed Hadley. This serial gives you the opportunity to see an interesting western, where you will only discover the real villain, Don del Oro, at its end. The serial also has good performance of various actors of movies B like Ed Cobb, ex$\_$ Tarzan Jim Pierce, C. Montague Shaw, eternal villains like John Merton and {\color{red}waste} ({\color{blue}charles}) King, and a very good performance of Hadley as Zorro. He was quick, smart, used well his whip and sword, and his voice was the best for any Zorro.\\
\hline
Negative&Positive&Well it certainly stunned me $\_$ I can not believe that someone made another Australian film thats even more boring than Somersault. The story is implausible, the characters, with the exception of Friels and Mailmans characters, are {\color{red}fun} ({\color{blue}unlikeable}) and wooden, Tom Long possesses a VAST array of facial expressions: happy and not happy, and the sex scenes, which could have been very confronting and disturbingly erotic, would have been at home in a low$\_$budget porno flick.This is the first movie I have seen in 30 years of cinema$\_$going that has had me on the edge of my seat....ready to get up and leave.The best thing about this movie is the promotional poster.\\
\hline
Negative&Positive&Stumbled over this film on Amazon.com. Had never heard of its release but the three reviews gave it five stars and rave reviews so being a lover of German movies I bought a copy...Have to say that I was {\color{red}amazing} ({\color{blue}not}) impressed. The production values are cheap, the story is derivative, the characters are less than engaging and for a comedy it is surprisingly short on laughs.I wanted to like this but I just found it lackluster and dull. Or maybe I expected more of independent German cinema than a gay spin on The Full Monty and a cast of stereotypes.There are bits in the film that make no sense at all, like one of the Leather Bears trying to get Ecki in a sling $\_$$\_$like hed even look at him twice? Or the vengeful ex$\_$wife turning up at the match but ending up cheering for her estranged gay husband? Bunkum is not the word! Well, at least it explains the movies UK title, I suppose...\\
\hline
Positive&Negative&Joan Cusack steals the show! The premise is good, the plot line {\color{red}script} ({\color{blue}interesting}) and the screenplay was OK. A tad too simplistic in that a coming$\_$out story of a gay man was so positive when it is usually not quite$\_$so$\_$positive. Then again, it IS fiction. :) All in all an entertaining romp.One thing I noticed was the inside$\_$joke aspect. Since the target$\_$audience probably was straight, they may not get the gay stuff in context with the story. Kevin Kline showed a facet of his acting prowess that screenwriters sometimes dont take in consideration when suggesting Kline for a part.This one hit the mark.\\
\hline
\hline
\caption{Examples of successful Gumbel Attacks on IMDB with Word-CNN. Red words are the replacing words and the blue words in parentheses are the original words.}
\label{tab:demo_IMDB_gumbel}
\end{longtable}

\newpage
\subsection{Visualization on AG's News with Char-CNN}

\begin{longtable}{||c|c|p{12cm}||}
\caption{AG's News Adversarial Examples From Greedy Attack}\\
 \hline
Class &  Perturbed Class & Purturbed Texts \\ [0.5ex] 
\hline\hline
Sports&Sci \& Tech&DEFOE DRIVES SPURS HOMEJermain Defoe underlined his claims for an improved contract as he inspired Tottenham to a 2$\_$0 win against 10$\_$man Middlesbrough. New {\color{red}s}{\color{red}x}{\color{red}$\backslash$}{\color{red}$\backslash$}{\color{red}$\backslash$} Martin Jol, who secured his first win in charge, may have been helped\\
\hline
World&Sci \& Tech&Vietnam Opens Bunker Used by Ho Chi Minh (AP)AP $\_$ Behind thick concrete walls and iron doors, Ho Chi Minh and other top Vietnamese le{\color{red}c}{\color{red}s}{\color{red}$\backslash$}rs hid in secret underground tunnels during U.S. B$\_$52 {\color{red}$\backslash$}ombing raids to plot key mi{\color{red}b}itary strategies that led to America's defeat in the Vietnam War.\\
\hline
Business&Sci \& Tech&Saks announces store closingsBIRMINGHAM, Ala. They're closing eight Saks Fifth Avenue stores and three Off Fifth outlet stores. Saks I{\color{red}p}{\color{red}$\backslash$}o{\color{red}$\backslash$}{\color{red}$\backslash$}o{\color{red}$\backslash$}ated says shutting down weaker stores will allow the company to focus on its more ``productive'' locations and further strengthen its brand.\\
\hline
Sci \& Tech&Business&Oracle Moves To Monthly Patch ScheduleAn alert posted on the company's {\color{red}y}{\color{red})}{\color{red}c} {\color{red}t}ite outlined the patches that should be posted to fix numerous security holes in a number of a{\color{red}i}plications.\\
\hline
World&Business&Jeanne death toll over 600 in {\color{red}$\backslash$}aiti(09/21/04)$\_$$\_$ The death toll keeps ris{\color{red}u}ng in Haiti. Officials say at least 622 people have been {\color{red}$\backslash$}il{\color{red}x}e{\color{red}c} by Hurricane Jeanne. Jeanne was downgraded to a tropical depression.\\
\hline
Sci \& Tech&Sports&Sony NW$\_$E95 and NW$\_$E99 Network WalkmanSony Europe has launched two tiny 512MB and 1G{\color{red}c} {\color{red}l}{\color{red}h}{\color{red}r} players, the NW$\_$E95 and NW$\_$E99 Network Walkman. Both play MP3 (Sony has officially bit the mp3 bullet) and ATRAC3plus compressed files and have a small blue backlit LCD s{\color{red}k}reen.\\
\hline
Business&World&Howard Stern moves radio show to S{\color{red}k}riusSho{\color{red}p}k jock Howard Stern announced Wednesday he's taking his radio show off the public airwaves and over to Sirius sat{\color{red}i}{\color{red}h}l{\color{red}h}te radio.\\
\hline
Sports&Sci \& Tech&England seek first one$\_$day title against surprise package WindiesLONDON: England have never won a major international limited$\_$overs title while West Indies Wor{\color{red}$\backslash$}{\color{red}$\backslash$} {\color{red}$\backslash$}{\color{red}$\backslash$}{\color{red}x} glory days date back to 1975 and 1979.\\
\hline
World&Sports&Rain threatens triangular final (A{\color{red}p}{\color{red}c}{\color{red}l}{\color{red}p}{\color{red}6}P $\_$ Organisers were left banking on the Dutch weather to spare Saturday's final of the triangular cricket tournament after deciding against altering the fixture schedule in a bid to beat the rain that has marred this warm$\_$up event for next month's ICC Champions Trophy in England.\\
\hline
Business&Sci \& Tech&Legal Loophole Inflates P{\color{red}$\backslash$}o{\color{red}$\backslash$}{\color{red}$\backslash$}ts in Student LoansThe White House could have closed a loophole through which stud{\color{red}$\backslash$}nt loan companies are billing the federal government nearly a billion dol{\color{red}i}ars, but chose not to.\\
\hline
World&Sci \& Tech&Video Shows March Madrid {\color{red}$\backslash$}o{\color{red}$\backslash$}{\color{red}$\backslash$}ingA ball of fire erupts from a train car, smothering com{\color{red}p}uters with smoke and littering the platform with bodies and staining it with blood in a chilling security$\_$camera videotape of the March 11 train {\color{red}f}ombings broadcast Tuesday by a Spanish station.\\
\hline
Sports&Sci \& Tech&The Newest Hope ; Marriage of Necessity Just Might Work OutNEW YORK $\_$ The TV lights were on, the cameras rolled and the symphony {\color{red}$\backslash$}f cameras flashing in his face blinded Ped{\color{red}$\backslash$}{\color{red}$\backslash$} {\color{red}$\backslash$}a{\color{red}$\backslash$}tinez $\_$ but not for long.\\
\hline
World&Business&Iran given nuclear deadlinethe (International Ato{\color{red}r}ic Energy {\color{red}x}gency) board of governors,'' he said. ``A proposal will be put to them.''. produce fuel for {\color{red}k}{\color{red}s}c{\color{red}c}ear weapons $\_$$\_$ but Tehran rejected the demand as illegal.\\
\hline
Sci \& Tech&World&No Sign Yet of Predicted Big {\color{red}l}a{\color{red}m}ifornia Earthquake (Reuters)Reuters $\_$ The clock is running out on a$\backslash$high{\color{red}e}{\color{red}n} {\color{red}j}ublicized prediction that a major earthquake will rip$\backslash$through Southern California by Sunday.\\
\hline
Sci \& Tech&Sports&Summary Box: Breast Cancer Surgery Refined (AP)A{\color{red}6} $\_$ NEW APPROACH: A st{\color{red}g}{\color{red}w}y {\color{red}n}ays that removing just one to three key lymph nodes can spare women lifelong arm problems and reliably indicate whether breast cancer has spread.\\
\hline
Business&World&Tokyo Stocks Flat, Focus on Data TOKYO (Reute{\color{red}l}s) $\_$ Japanese {\color{red}y}to{\color{red}k}{\color{red}$\_$}s were flat in mid$\_$morning  trade on Monday with confidence in the domestic economic  outlook failing to offset pro{\color{red}o}it$\_$taking that hit recent gainers  such as insurers and real estate stocks.\\
\hline
Sci \& Tech&Business&Dell Takes Another Cut at Blade Market``The biggest danger to HP an{\color{red}\textdollar} IBM is a price war,'' said John Enck of Gartner. ``Blades are still premium$\_${\color{red}b}riced pr{\color{red}\textdollar}duc{\color{red}k}{\color{red}\textdollar} from IBM and HP.\\
\hline
Business&World&Putin ready to probe other o{\color{red}f}l {\color{red}b}o{\color{red}d}{\color{red}1}aniesRussian President Vladimir Putin is ready to go after other oil companies the way he has hammered Yukos, a top Kremlin official has said.\\
\hline
Sports&World&ROUNDUP: Franchitti overcomes pit mishap for IRL win{\color{red}h}OUN{\color{red}i}AIN, Colo. $\_$$\_$ Dario Franchitti shook off a potentially dangerous pit mishap to win t{\color{red}c}{\color{red}k} IRL{\color{red}m}s Honda 225 Sunday at Pikes Peak International Raceway.\\
\hline
Business&World&Boeing Offers 777 Cargo FreighterBoei{\color{red}u}g {\color{red}a}{\color{red}f}{\color{red}b} on {\color{red}l}onday said it is offering a 777 cargo model that will be the largest and farthest$\_$flying twin$\_$engine freighter. The Boeing 777 freighter is scheduled to enter service in the fourth quarter of 2008.\\
\hline
Business&Sci \& Tech&VZ Wireless Slams National 411 DirectoryWASHINGTON $\_$$\_$ Ver{\color{red}s}zon Wireless, the n{\color{red}e}tion's lar{\color{red}v}{\color{red}$\backslash$}{\color{red}$\backslash$}t wireless carrier, clashed with other cellular carriers on Tuesday, telling a US Senate committee that a proposal for a national wireless telephone directory is a ``terrible idea'' and that the proposal\\
\hline
World&Business&Insurgents target Green ZoneInsurgents exploded two car bombs at the gates of the main US$\_$I{\color{red}p}{\color{red}x}{\color{red}x}i headquarters in {\color{red}c}{\color{red}x}ghdad and near major hotels Monday, killing at least 21 people and wounding 96.\\
\hline
Business&Sci \& Tech&Lakshmi Mittal is Britain's richest manLondon: NRI business tycoon Lakshmi N Mittal, who is set to control the world{\color{red}p}s {\color{red}$\backslash$}a{\color{red}c}{\color{red}$\backslash$}{\color{red}$\backslash$}st steelmaker, has emerged as the richest man in Britain.\\
\hline
Sci \& Tech&Sports&Game sparks sales frenzyGAMES stor{\color{red}1}s opened at midnight to meet demand for the {\color{red}k}atest ve{\color{red}s}{\color{red}r}ion of the controversial Great Theft Auto. There were queues outside shops around {\color{red}j}erseyside with people anxious  to get their hands on copies of GTA San Andreas.\\
\hline
World&Sci \& Tech&After a decade of silence, Cambodia's cinema enjoys resurgence (AFP)AFP $\_$ After a decade of silence, Cam{\color{red}$\backslash$}o{\color{red}$\backslash$}{\color{red}$\backslash$}{\color{red}.}{\color{red}$\backslash$}s movie industry is enjoying a boom due to a dash of Hollywood attention, burgeoning nationalism and cheaper production costs, industry insiders say.\\
\hline
Sports&World&Jerkens makes right call with Society SelectionTraine{\color{red}k} Alle{\color{red}a} Jerkens hemmed and hawed this past week over running {\color{red}h}ociety Selection in Saturda{\color{red}q}'s Grad{\color{red}d} 1 Alabama at Saratoga.\\
\hline
World&Sci \& Tech&French Hostage Transfer Sparks Release Hopes PARIS (Reuters) $\_$ Hopes of a swift end to the French  {\color{red}r}ostage crisis rose early Friday, after the Le Figaro newspa{\color{red}m}er  that employs one of the two captives said the men were now  being held by I{\color{red}$\backslash$}{\color{red}$\backslash$}{\color{red}x}i guerrillas willing to negotiate their  release.\\
\hline
\hline
\caption{Examples of successful greedy attacks on AG's News with Char-CNN. Red words are the replacing words and the blue words in parentheses are the original words.}
\label{tab:demo_ag_greedy}
\end{longtable}

\begin{longtable}{||c|c|p{12cm}||}
\caption{AG's News Adversarial Examples From Gumbel Attack}\\
 \hline
Class &  Perturbed Class & Purturbed Texts \\ [0.5ex] 
\hline\hline
Sports&Sci \& Tech&Anniversary remembered on game dayWhen the attacks came on S{\color{red}x}pt. 11, 2001, Tom O'Brien, if only for a moment, stopped being Boston College's {\color{red}$\backslash$}{\color{red}$\backslash$}{\color{red}$\backslash$}c{\color{red}$\backslash$}. On that day, as the World Trade Center and Pentagon smoldered and the world stood still, O'Brien was a Navy man.\\
\hline
Sports&Sci \& Tech&Prem Preview: Everton v FulhamA favourite with the crowd during his time at Goodison Park, many fans will n{\color{red}$\backslash$}t fo{\color{red}g}get R{\color{red}x}dzi{\color{red}$\backslash$}skis comments prior to his 1.75 m move to the Londoners du{\color{red}o}ing the summer.\\
\hline
Sports&Sci \& Tech&In OT, first$\_$minute men lift UMassSteph{\color{red}c}n Werner and the Minutemen weren't easily discouraged in their Hockey East opener las{\color{red}b} {\color{red}$\backslash$}{\color{red}o}g{\color{red}$\backslash$}t.\\
\hline
Business&Sci \& Tech&Plan to ease sale of Abbey sharesAbbey National s{\color{red}j}ar{\color{red}x}holders will no longer need to fill in complex Spanish t{\color{red}$\backslash$}x fo{\color{red}x}ms if BSCH's {\color{red}r}id to buy the UK firm succeeds.\\
\hline
Sci \& Tech&Business&Web domains approved for Posties, travelThe Int{\color{red}a}{\color{red}x}{\color{red}e}e{\color{red}r} Corporation for Assigned Names and Numbers (ICANN) has approved two new sponsored Interne{\color{red}d} domains, .post and .travel, specifically for the post and travel industries.\\
\hline
World&Business&Philippine Floods Kill HundredsMore than 300 pe{\color{red}c}{\color{red}$\backslash$}{\color{red}x}e died after flash floods and landslides devastated three coastal towns and left swathes of the northern Ph{\color{red}g}lippines under water on Tuesday.\\
\hline
Sports&World&Gordon favored in 'chase'LOUDEN, NH {\color{red}k}$\_$ Right now, things are going Jeff Gordon's way. That should enhance his chances of winning a fifth NASCAR {\color{red}$\_$}h{\color{red})}{\color{red}i}pionship.\\
\hline
Sci \& Tech&Business&IBM to use dual$\_$core OpteronBig Blue will use AMD's chip in a hi{\color{red}t}h$\_$performance {\color{red}a}er{\color{red}g}e{\color{red}b} but isn't yet planning a general$\_$purpose Opteron {\color{red}h}ystem.\\
\hline
World&Sci \& Tech&Darfur Peace Talks Struggle for Survival ABUJA (Reuters) $\_$ Peace talks between Sudan's government  and Darf{\color{red}$\backslash$}{\color{red}$\backslash$} reb{\color{red}$\backslash$}{\color{red}y}s st{\color{red}$\backslash$}uggled for survival after one of the two  rebel groups said on Wednesday the negotiations had collapsed  but left open the chance of resumption.\\
\hline
World&Business&Medicare Premiums to Rise Record 17 Pct.WASHINGTON {\color{red}$\backslash$} Medicare prem{\color{red};}ums for doctor visits are going up a record \textdollar11.60 a month next year. The Bush administration says the increase reflects a strengthened Medicare, while {\color{red}x}emocrats complain that seniors are being unfairly socked{\color{red}$\backslash$}.{\color{red}$\backslash$}\\
\hline
World&Business&In Chile, pace of justice quickensA judge has ruled that Gen. Augusto {\color{red}y}{\color{red}k}n{\color{red}x}{\color{red}$\backslash$}het stand trial for his alleged involvement in state{\color{red}h}sponsored torture.\\
\hline
Sports&Sci \& Tech&No. 3 Miami Stops No. 18 Louisville 41{\color{red}i}38 {\color{red}q}A{\color{red}i}{\color{red}q}AP $\_$ The Louisville Cardinals drew a flag for excessive celebration in the second quarter, and another in the third. Against Miami, the displays of j{\color{red}$\_$}bilation were premature. Led by Brock Berlin and Devin Hester, the third$\_$ranked Hurricanes erased a 17$\_$point deficit over the final 20 minutes and came from behind twice in the fourth quarter to beat No. 18 Louisville 41$\_$38 Thursday night.\\
\hline
World&Business&Israel scrambled warplanes 'in case of hijacking threat' Is{\color{red}$\backslash$}{\color{red}d}e{\color{red}f}i warplanes scrambled as soon as news broke of the Taba {\color{red}$\backslash$}o{\color{red}s}bings. Military sources would not elaborate but analysts suggested the most likely reason was to intercept any hijacked\\
\hline
Sports&World&Donald runs into Ryder formLuke Donal{\color{red}k} says his win in {\color{red}e}he Europea{\color{red}c} M{\color{red}$\_$}st{\color{red}y}rs on Sunday bodes well for his upcoming Ryder Cup debut. Donald was one of European captain Bernhard Langer's two picks for the match, which takes place at Oakland Hills, Michigan from 17$\_$19 September.\\
\hline
Business&Sci \& Tech&AMP shrugs off British debacleAustralian insurer AMP returned to the black in the first half of the year with net {\color{red}$\backslash$}ro{\color{red}x}{\color{red}$\backslash$}{\color{red}$\backslash$}s of A$\backslash${\color{red}$\backslash$}378m (150m) after a disastrous foray into Britain pushed it A \textdollar2.16 billion into the red last year.\\
\hline
Sci \& Tech&Business&New RIAA File$\_$swapping Suits Target StudentsFletcher writes ``The Reco{\color{red}.}{\color{red}p}ing Industry Associa{\color{red}n}ion of America filed another round of lawsuits against alleged file$\_$swappers, including students on 13 unive{\color{red}q}sity camp{\color{red};}ses.\\
\hline
World&Sci \& Tech&Soldiers face Abu Ghraib hearingsFour US sold{\color{red}s}ers charged with abusing {\color{red}$\backslash$}{\color{red}h}{\color{red}$\backslash$}{\color{red}x}i prisoners  are set to face pre$\_$trial hearings in Germany.\\
\hline
Sci \& Tech&Business&Former Dot$\_$Com Commerce One Eyes ClosureCommerce One Inc., an Inte{\color{red}\textdollar}{\color{red}a}{\color{red}i}{\color{red}r} soft{\color{red}m}are maker valued at \textdollar20 billion at the peak of dot$\_$com mania, is poised to go out of business as a pauper.\\
\hline
Sci \& Tech&Sports&Fire Pit Dated to Be Over 50,000 Years Old (A{\color{red}1})AP $\_$ In the growing debate about when people first appeared on this continent, a leading archae{\color{red}j}l{\color{red}f}{\color{red}h}ist said Wednesday he has di{\color{red}i}covered what could be sooty evidence of human occupation in North America tens of thousands of years earlier than is commonly believed.\\
\hline
Business&Sports&Enron of Kansas' Trial BeginsIn the recent annals of {\color{red}1}{\color{red}j}{\color{red}u}{\color{red}b}o{\color{red}3}ate fraud, the names Enron, Tyco and WorldCom ring the loudest. But for residents of Topeka, Kan., the former leaders of the local utility company have become just as infamous.\\
\hline
Business&Sci \& Tech&Schering$\_$Plough and Bayer form strategic all{\color{red}$\backslash$}a{\color{red}.}ceSchering$\_$Plough Corporation has announced that it has entered into a strategic agreement with Bayer designed to maximize the companies' pharmace{\color{red}p}tical resources while maintaining each c{\color{red}y}mpan{\color{red}n}'s own strategic interests.\\
\hline
Sci \& Tech&World&TV aims for prime time in digital homeNew standard use{\color{red}q} {\color{red}d}{\color{red}b}{\color{red}n}{\color{red}m}based protocols to let televisions control other devices in a home.\\
\hline
Business&World&EU, US Talks On Aircraft Aid GroundedUS and EU negotiators traded arguments on Thursday over state aid for aircraft rivals {\color{red}j}{\color{red}$\_$}{\color{red}i}{\color{red}u}{\color{red}n}s and Boeing, but wound up no closer on a sensitive issue that has gathered steam in the run up to the US presidential election.\\
\hline
Sports&Business&Murphy Acquitted of Sexual Abuse ChargesBask{\color{red}f}t{\color{red}a}{\color{red}x}ll Hall of Fame member Calvin Murphy, left, sits in a courtroom as he waits for the start of closing arguments in his trial Monday, Dec. 6, 2004, in Houston.\\
\hline
Business&Sci \& Tech&Oil Near \textdollar50 on Supply Fears in NigeriaOil prices {\color{red}n}ose to record highs Monday near \textdollar50 a barrel for {\color{red}i}{\color{red}j} crude as Nigeria emerged as the latest focus for worries about supply in an already tight worldwide ene{\color{red}b}gy ma{\color{red}$\backslash$}ket.\\
\hline
Sci \& Tech&Business&Tiny telescope's big di{\color{red}\textdollar}cove{\color{red}m}y opens new doorsWASHINGTON $\_$ A tiny telescope has spotted a giant planet circling a faraway star, using a technique that could open a new phase of planetary di{\color{red}o}{\color{red}v}overy.\\
\hline
Business&World&Treasury Prices Crawl Higher Before Data NEW YORK (Reuters) {\color{red}d} {\color{red}p}re{\color{red}u}{\color{red}j}uries prices crawled ahead on  Tuesday as a hesitant market awaited the latest reading on  consu{\color{red}s}er sentiment and an auction of new U.S. government debt.\\
\hline
World&Business&Congo Ex$\_$Rebel Group Pulls Out of Government (Reuters)Reuters $\_$ The former main {\color{red}x}{\color{red}x}{\color{red}$\backslash$}e{\color{red}x} group during$\backslash$Congo's civil war pulled out of a power$\_$sharing transitional$\backslash$government on Monday, dealing a major blow to the country's$\backslash$already fragile pe{\color{red}\textdollar}ce process.\\
\hline
World&Sports&Flight Diverted to UK After {\color{red}v}o{\color{red}$\backslash$}{\color{red}$\backslash$} ThreatA Singapore Airlines passenger jet from Frankfurt to New York was diverted to Manchester A{\color{red}w}{\color{red}a}port in northern England on Monday after a bomb threat that police said may have been a hoax.\\
\hline
Sci \& Tech&Business&NASA 'scramjet' makes historic flight off CaliforniaLOS ANGELES {\color{red}v}A{\color{red}h}{\color{red}1}'s unmanned ``{\color{red}u}cra{\color{red}/}jet'' proved it's small but it's fast $\_$$\_$ in a record$\_$breaking demonstration above the Pacific Ocean.\\
\hline
\hline
\caption{Examples of successful Gumbel Attacks on AG's News with Char-CNN. Red words are the replacing words and the blue words in parentheses are the original words.}
\label{tab:demo_ag_gumbel}
\end{longtable}

\newpage
\subsection{Visualization on Yahoo! Answers with LSTM}

\begin{longtable}{||c|c|p{10cm}||}
\caption{Yahoo! Answers Adversarial Examples From Greedy Attack}\\
 \hline
Class &  Perturbed Class & Purturbed Texts \\ [0.5ex] 
\hline\hline
Sports&Society, Culture&are {\color{red}jesus} ({\color{blue}wwe}) {\color{red}jesus} ({\color{blue}wrestlers}) are really getting hurts while fighting yes totally\\
\hline
Computers, Internet&Business, Finance&where in the us can i find a good sales person for a {\color{red}bush} ({\color{blue}software}) engineering firm any web sites looking for a qualified cold caller in the us that how to generate outsourcing business in the {\color{red}doctor} ({\color{blue}software}) industry try http www monster com\\
\hline
Computers, Internet&Health&i only got 12 free space i need get some stuff off so i can have more space so i can {\color{red}doctor} ({\color{blue}defrag}) can someone help go to n nhttp www {\color{red}doctor} ({\color{blue}ccleaner}) com\\
\hline
Education, Reference&Politics, Government&is it no {\color{red}bush} ({\color{blue}one}) or noone or are both correct no {\color{red}bush} ({\color{blue}one}) is correct\\
\hline
Politics, Government&Society, Culture&what does the {\color{red}jesus} ({\color{blue}aclu}) think it is doing other than being a i mean honestly free speech is important but people also have to have decency they are helping to strip the {\color{red}jesus} ({\color{blue}nation}) of our the values and that make us americans they are ensuring that no one is judged based on their actions that anything and everything goes n nthey used to protect americans right to free speech but now they are so far left they make the 9th circus court of appeals appear right wing\\
\hline
Education, Reference&Entertainment, Music&what is the role of radio tv as a public {\color{red}song} ({\color{blue}relations}) channel i need full information on the above topic to spread the {\color{red}music} ({\color{blue}liberal}) left wing agenda\\
\hline
Family, Relationships&Health&im {\color{red}weight} ({\color{blue}bored}) so whats a good {\color{red}weight} ({\color{blue}prank}) so i can do it on my friends go to their house and dump all the shampoo outta the bottle and replace it with yogurt yup i always wanted to do that let me know how it works out haha\\
\hline
Politics, Government&Entertainment, Music&is g {\color{red}music} ({\color{blue}bush}) an idiot it seems he picks stupid people to work for him and he offers no for any thing picks stupid people to work for him would you like to add more info to this what people are you talking about this would make it alot for {\color{red}song} ({\color{blue}us}) to talk about if we are all on the same page\\
\hline
Family, Relationships&Society, Culture&what questions should i ask my {\color{red}jesus} ({\color{blue}mate}) before getting {\color{red}bush} ({\color{blue}married}) let's see are you willing to spend the rest of your life with me what is your view on divorce have you ever been convicted of a felony do you have twins in the family does anyone in your family have any major health problems how do feel about children how would you raise a child are you happy with your career my list can go on sit down with your mate and talk about your ideas and concerns before you get married it's a great way to see his her point of view best of luck\\
\hline
Politics, Government&Society, Culture&do you agree w the {\color{red}jesus} ({\color{blue}aclu}) and there opinion with the seperation of church and state they say that according to the that they are suppose to be seperate however if you go back in history clearly the founding fathers promoted thomas jefferson himself the one who wrote the seperation of church state made it to promote to the indians one year later and use bibles as national school books why do you think the aclu is trying to change the meaning of our {\color{red}jesus} ({\color{blue}constitution}) and do you think people even do research to understand what jefferson meant before holding an opinion they are just a bunch of lawyers who enjoy screwing with the law just because they can if they dedicated their time to criminals it would be a better world to live in\\
\hline
Business, Finance&Politics, Government&can a annuity be rolled over to an {\color{red}bush} ({\color{blue}ira}) cd account at a bank without paying income tax it depends on the status of the tax annuity if the is an ira b 401 k or some sort of account that you took a tax deduction on your tax return the year s you made contributions then you can roll it to an ira cd at the bank look carefully at your statement from the and see if it says ira or 401 k or b or qualified account close to where your name or account number is located if you added money to this account thru a payroll deduction plan it probably is an account that can be rolled to the ira cd na can also be what is called a non qualified account meaning that you put after tax dollars into the account these accounts cannot be rolled over to a cd nit is often times hard to get the company or their representative to help you with you questions about moving money out of their company however the bank where you want to move the money to should have a representative that is more than willing to help you figure out how to move money to their bank take your statement to the bank and ask for their help i disagree with the previous answer about getting help from the {\color{red}bush} ({\color{blue}company})\\
\hline
Entertainment, Music&Sports&what are some really good dave matthews {\color{red}cup} ({\color{blue}band}) {\color{red}cup} ({\color{blue}songs}) ants marching n marching though would probably be my favorite or the first one i would recommend\\
\hline
Entertainment, Music&Sports&don't everyone answer at once but i think the damned were the best real punk {\color{red}cup} ({\color{blue}band}) who here agrees green day {\color{red}cup} ({\color{blue}fans}) need not answer please wot ya reckon to the\\ 
\hline
\hline
\caption{Examples of successful Greedy Attacks on Yahoo!\ Answers with LSTM. Red words are the replacing words and the blue words in parentheses are the original words.}
\label{tab:demo_yahoo_greedy}
\end{longtable} 

\begin{longtable}{||c|c|p{10cm}||}
\caption{Yahoo! Answers Adversarial Examples From Gumbel Attack}\\
 \hline
Class &  Perturbed Class & Purturbed Texts \\ [0.5ex] 
\hline\hline
Family, Relationships&Politics, Government&my freind that is a boy {\color{red}bush} ({\color{blue}jacks}) off in front of me how can i make hi stop {\color{red}bush} ({\color{blue}nasty}) the same thing i do when my dog tries to hump someones leg always carry around a water bottle and squirt him with it when he starts\\
\hline
Business, Finance&Sports&trying to find lost {\color{red}cup} ({\color{blue}401k}) left at warren ford 10 or 11 yrs ago how can i find com can't remember com name who had 401k need to find out com name warren ford used for our 401k go to the courthouse in that town and they maybe able to help you with this they maybe able to tell you if a bank was involved with their business and then they may know more but wouldnt the state of florida have that {\color{red}cup} ({\color{blue}info}) also\\
\hline
Politics, Government&Sports&how many terms has franklin served {\color{red}cup} ({\color{blue}usa}) which party was he in he was elected to four terms however he died a few months into his forth term therefore he was president for a little over 12 years he was a {\color{red}cup} ({\color{blue}democrat})\\
\hline
Politics, Government&Sports&who is the current {\color{red}cup} ({\color{blue}president}) of south africa n non june 2 1999 the deputy {\color{red}cup} ({\color{blue}president}) of south africa and leader of the african national congress was elected president in a landslide having already assumed many of governing responsibilities shortly after mandela won south first democratic election in 1994\\
\hline
Computers, Internet&Politics, Government&i can not access some pictures off my {\color{red}bush} ({\color{blue}email}) there is 6 boxes w an x in the middle do i need a spec program you need to have an internet connection to see them when you have pictures pasted to an email they are being shown from the website they are hosted at if you don't have an internet connection at the time you are looking at the email the images won't show up n to the internet and then refresh the email and they should work as long as they are still on the hosting {\color{red}bush} ({\color{blue}server}) where they were copied from\\
\hline
Politics, Government&Society, Culture&does nancy have a lot of nerve or what actively in the slaughter of the intelligence agencies of {\color{red}jesus} ({\color{blue}america}) and then has the nerve to criticize them as a member of the 9 11 commission n http clerk house gov 1997 n n1998 http clerk house gov 1998 n http clerk house gov 1999 she's the leader of the party n nhot damn can't wait for {\color{red}jesus} ({\color{blue}2008})\\
\hline
Computers, Internet&Politics, Government&does anyone know a good website for {\color{red}bush} ({\color{blue}photoshop}) {\color{red}bush} ({\color{blue}tutorials}) http www org web photoshop tutorials\\
\hline
Business, Finance&Politics, Government&how do you obtain a patent for your customized designs find a {\color{red}bush} ({\color{blue}patent}) attorney who can your ideas they do a patent search and then submit the information the the office of patent and trademarks in md n nmy mother invented the laundry us patent 4 good {\color{red}bush} ({\color{blue}luck})\\
\hline
Education, Reference&Computers, Internet&what is the of abu u a e hello i have ask this question before and the one and only reply was wrong so i decided to ask this again i don't care about points so plz tell me what is the of abu u a e united {\color{red}windows} ({\color{blue}arab}) {\color{red}windows} ({\color{blue}emirates}) postal zip code\\
\hline
Education, Reference&Politics, Government&why did john lee {\color{red}bush} ({\color{blue}love}) invent the pencil john lee {\color{red}bush} ({\color{blue}love}) to pencils\\
\hline
Politics, Government&Entertainment, Music&do you feel jack got off too easy was his {\color{red}song} ({\color{blue}plea}) bargain fair to the usa and humanity it is fair if he delivers some of the bribe takers to the justice system he can be very to the {\color{red}song} ({\color{blue}government}) in this way there are bigger fish in the pond then jack believe me\\
\hline
Family, Relationships&Sports&anybody ever felt relieved with breaking up with a {\color{red}cup} ({\color{blue}lover}) yes a {\color{red}cup} ({\color{blue}boy}) who didn't want to grow up and didn't care about anything or anyone except himself and and video games\\
\hline
Computers, Internet&Society, Culture&what is the best way to learn i have solid exp in core {\color{red}jesus} ({\color{blue}java}) and jsp but i face problem 2 learn {\color{red}jesus} ({\color{blue}java}) plz help how i learn it step by step what are data structures you'll have to rephrase your question so people know what you are talking about\\
\hline
Computers, Internet&Politics, Government&what are legacy {\color{red}bush} ({\color{blue}data}) {\color{red}bush} ({\color{blue}and}) what are the options for their incorporation into an e commerce system you need to search all of that and its not easy question for an ordinary person it's either theory or a real practical concept but i have another for you because i think you are a head of me in e commerce what is switching cost and transaction cost are they different from each other or show me their similarities e mail me if you have the answer please yahoo com\\
\hline
Computers, Internet&Politics, Government&how can i {\color{red}bush} ({\color{blue}bypass}) {\color{red}bush} ({\color{blue}parental}) controls my dad is gone for a month and he accidentally blocked me from any sites something tells me it wasn't accidental n nnice try though\\
\hline
\hline
\caption{Examples of successful Gumbel Attacks on Yahoo!\ Answers with LSTM. Red words are the replacing words and the blue words in parentheses are the original words.}
\label{tab:demo_yahoo_gumbel}
\end{longtable} 
\end{document}